\documentclass[12pt,twoside]{article}
  \newdimen\paravsp  \paravsp=1.3ex
\usepackage{amsthm}
\usepackage[sumlimits]{amsmath}
\usepackage{amssymb}
\usepackage{algorithm}
\usepackage{algorithmic}
\usepackage{txfonts}
\usepackage{graphicx}
\usepackage{subfigure}
\usepackage[all]{xy}
\usepackage{url}

\newtheorem{defn}{Definition}
\newtheorem{lem}{Lemma}
\newtheorem{thm}{Theorem}

\newtheorem{proposition}{Proposition}

\theoremstyle{definition}

\def\citeA{\cite}
\newenvironment{keywords}{\centerline{\bf\small
Keywords}\begin{quote}\small}{\par\end{quote}\vskip 1ex}

\def\paradot#1{\paragraph{#1.}}

\newcommand{\cbar}{\,|\,}
\newcommand{\cdbar}{\,|\,}
\newcommand{\dens}{{\it Density}}
\newcommand{\cA}{{\cal A}}
\newcommand{\cO}{{\cal O}}
\newcommand{\cR}{{\cal R}}
\newcommand{\cX}{{\cal X}}
\newcommand{\searchalg}{{$\rho$UCT}}

\newcommand{\bstr}[1]{\llbracket #1 \rrbracket}

\newcommand{\cP}{{\cal P}}
\newcommand{\agent}{\text{MC-AIXI({\sc fac-ctw})}}
\newcommand{\predictor}{\text{\sc FAC-CTW}}

\begin{document}

\title{\vspace{-4ex}\bf\LARGE\hrule height5pt \vskip 4mm
A Monte-Carlo AIXI Approximation
\vskip 4mm \hrule height2pt}
\date{December 2010}
\def\name{\bf\small}
\def\email{\hfill\sc}
\def\addr#1{\small\it #1\hfill}
\def\AuthorAND{\\[-0.3ex]}
\def\Anl{\\[-1ex]}

\author{
  \name Joel Veness \email joelv@cse.unsw.edu.au \Anl
  \addr{University of New South Wales and National ICT Australia}
 \AuthorAND
  \name Kee Siong Ng \email keesiong.ng@gmail.com \Anl
  \addr{The Australian National University}
 \AuthorAND
   \name Marcus Hutter \email marcus.hutter@anu.edu.au \Anl
   \addr{The Australian National University and National ICT Australia}
 \AuthorAND
   \name William Uther \email william.uther@nicta.com.au \Anl
   \addr{National ICT Australia and University of New South Wales}
 \AuthorAND
   \name David Silver \email davidstarsilver@googlemail.com \Anl
   \addr{Massachusetts Institute of Technology}
}

\maketitle

\vspace{-7ex}\begin{abstract}
This paper introduces a principled approach for the design of a scalable general reinforcement learning agent.
Our approach is based on a direct approximation of AIXI, a Bayesian optimality notion for general reinforcement learning agents.
Previously, it has been unclear whether the theory of AIXI could motivate the design of practical algorithms.
We answer this hitherto open question in the affirmative, by providing the first computationally feasible approximation to the AIXI agent.
To develop our approximation, we introduce a new Monte-Carlo Tree Search algorithm along with an agent-specific extension to the Context Tree Weighting algorithm.
Empirically, we present a set of encouraging results on a variety of stochastic and partially observable domains.
We conclude by proposing a number of directions for future research.
\def\contentsname{\centering\normalsize Contents}
{\parskip=-2.7ex\tableofcontents}
\end{abstract}

\vspace{-2ex}\begin{keywords}\vspace{-1.5ex}
Reinforcement Learning (RL);
Context Tree Weighting (CTW);
Monte-Carlo Tree Search (MCTS);
Upper Confidence bounds applied to Trees (UCT);
Partially Observable Markov Decision Process (POMDP);
Prediction Suffix Trees (PST).
\end{keywords}

\newpage
\section{Introduction}\label{Introduction}

Reinforcement Learning \cite{sutton-barto98} is a popular and influential paradigm for agents that learn from experience.
AIXI \cite{Hutter:04uaibook} is a Bayesian optimality notion for reinforcement learning agents in unknown environments.
This paper introduces and evaluates a practical reinforcement learning agent that is directly inspired by the AIXI theory.

\paradot{The General Reinforcement Learning Problem}
Consider an agent that exists within some unknown environment.
The agent interacts with the environment in cycles.
In each cycle, the agent executes an action and in turn receives an observation and a reward.
The only information available to the agent is its history of previous interactions.
The \emph{general reinforcement learning problem} is to construct an agent that, over time, collects as much reward as
possible from the (unknown) environment.

\paradot{The AIXI Agent}
The AIXI agent is a mathematical solution to the general reinforcement learning problem.
To achieve generality, the environment is assumed to be an unknown but computable function; i.e.\ the observations and rewards received by
the agent, given its past actions, can be computed by some program running on a Turing machine.
The AIXI agent results from a synthesis of two ideas:
\begin{enumerate}\itemsep1mm\parskip0mm
 \item the use of a finite-horizon expectimax operation from sequential decision theory for action selection; and
 \item an extension of Solomonoff's universal induction scheme \cite{solomonoff64} for future prediction in the agent context.
\end{enumerate}
More formally, let $U(q, a_1a_2\ldots a_n)$ denote the output of a universal Turing machine $U$ supplied with program $q$ and input
$a_1a_2\ldots a_n$, $m \in \mathbb{N}$ a finite lookahead horizon, and $\ell(q)$ the length in bits of program $q$.
The action picked by AIXI at time $t$, having executed actions $a_1a_2\ldots a_{t-1}$ and having received the sequence of observation-reward pairs
$o_1r_1o_2r_2\ldots o_{t-1}r_{t-1}$ from the environment, is given by:
\begin{equation}
\label{aixi_eq}
a_t^* = \arg\max\limits_{a_t}\sum\limits_{o_t r_t} \dots \max\limits_{a_{t+m}}\sum\limits_{o_{t+m} r_{t+m}}[r_t + \dots + r_{t+m}]
\sum\limits_{q:U(q,a_1...a_{t+m})=o_1 r_1 ... o_{t+m}r_{t+m}}2^{-\ell(q)}.
\end{equation}
Intuitively, the agent considers the sum of the total reward over all possible futures up to $m$ steps ahead,
weighs each of them by the complexity of programs consistent with the agent's past that can generate that future,
and then picks the action that maximises expected future rewards.
Equation (\ref{aixi_eq}) embodies in one line the major ideas of Bayes, Ockham, Epicurus, Turing, von Neumann, Bellman, Kolmogorov,
and Solomonoff.
The AIXI agent is rigorously shown by \citeA{Hutter:04uaibook} to be optimal in many different senses of the word.
In particular, the AIXI agent will rapidly learn an accurate model of the environment and proceed to act optimally to achieve its goal.

Accessible overviews of the AIXI agent have been given by both \citeA{LeggPHD08} and \citeA{hutter07topdown}.
A complete description of the agent can be found in \cite{Hutter:04uaibook}.

\paradot{AIXI as a Principle}
As the AIXI agent is only asymptotically computable, it is by no means an algorithmic solution to the general reinforcement learning problem.
Rather it is best understood as a Bayesian \emph{optimality notion} for decision making in general unknown environments.
As such, its role in general AI research should be viewed in, for example, the same way the minimax and empirical risk minimisation principles
are viewed in decision theory and statistical machine learning research.
These principles define what is optimal behaviour if computational complexity is not an issue, and can provide important theoretical guidance in the design of practical algorithms.
This paper demonstrates, for the first time, how a practical agent can be built from the AIXI theory.

\paradot{Approximating AIXI}
As can be seen in Equation~(\ref{aixi_eq}), there are two parts to AIXI.
The first is the expectimax search into the future which we will call {\em planning}.
The second is the use of a Bayesian mixture over Turing machines to predict future observations and rewards based on past experience; we will call that {\em learning}.
Both parts need to be approximated for computational tractability.
There are many different approaches one can try.
In this paper, we opted to use a generalised version of the UCT algorithm \cite{kocsis06} for
planning and a generalised version of the Context Tree Weighting algorithm \cite{ctw95} for learning.
This combination of ideas, together with the attendant theoretical and experimental results, form the main
contribution of this paper.

\paradot{Paper Organisation}
The paper is organised as follows.
Section \ref{sec:agent setting} introduces the notation and definitions we use to describe environments and accumulated agent experience, including the familiar notions of reward, policy and value functions for our setting.
Section \ref{sec:BayesAgents} describes a general Bayesian approach for learning a model of the environment.
Section \ref{sec:mcts} then presents a Monte-Carlo Tree Search procedure that we will use to approximate the expectimax operation in AIXI.
This is followed by a description of the Context Tree Weighting algorithm and how it can be generalised for use in the agent setting in Section \ref{sec:ctw}.
We put the two ideas together in Section \ref{sec:together} to form our AIXI approximation algorithm.
Experimental results are then presented in Sections \ref{sec:experiments}.
Section \ref{sec:discussion} provides a discussion of related work and the limitations of our current approach.
Section \ref{sec:future_work} highlights a number of areas for future investigation.

\section{The Agent Setting}\label{sec:agent setting}

This section introduces the notation and terminology we will use to describe strings of agent experience, the true underlying environment and the agent's model of the true environment.

\paradot{Notation}
A string $x_1x_2 \ldots x_n$ of length $n$ is denoted by $x_{1:n}$.
The prefix $x_{1:j}$ of $x_{1:n}$, $j\leq n$, is denoted by $x_{\leq j}$ or $x_{< j+1}$.
The notation generalises for blocks of symbols: e.g.\ $ax_{1:n}$ denotes $a_1x_1a_2x_2\ldots a_nx_n$ and $ax_{<j}$ denotes $a_1x_1 a_2x_2\ldots a_{j-1}x_{j-1}$.
The empty string is denoted by $\epsilon$.
The concatenation of two strings $s$ and $r$ is denoted by $sr$.

\paradot{Agent Setting}
The (finite) action, observation, and reward spaces are denoted by $\cA, \cO$, and $\cR$ respectively.
Also, ${\cal X}$ denotes the joint perception space $\cO \times \cR$.

\begin{defn}
A history $h$ is an element of $({\cal A} \times {\cal X})^* \cup ({\cal A} \times {\cal X})^* \times {\cal A}$.
\end{defn}

The following definition states that the environment takes the form of a probability distribution over possible observation-reward sequences
conditioned on actions taken by the agent.

\begin{defn}
\label{def:environment_model}
An environment $\rho$ is a sequence of conditional probability functions $\{ \rho_0, \rho_1, \rho_2, \dots \}$,
where $\rho_n \colon \mathcal{A}^n \rightarrow \dens \; (\mathcal{X}^n)$, that satisfies
\begin{equation}\label{eq:chronological}
\forall a_{1:n} \forall x_{<n}: \,\rho_{n-1}(x_{<n} \cbar a_{<n}) = \sum_{x_n \in {\cal X}}\, \rho_n(x_{1:n} \cbar a_{1:n}).
\end{equation}
In the base case, we have $\rho_0(\epsilon \cbar \epsilon) = 1$.
\end{defn}

Equation (\ref{eq:chronological}), called the chronological condition in \cite{Hutter:04uaibook}, captures the natural constraint that action $a_n$
has no effect on earlier perceptions $x_{<n}$.
For convenience, we drop the index $n$ in $\rho_n$ from here onwards.

Given an environment $\rho$, we define the predictive probability
\begin{equation}
 \rho( x_n \cbar ax_{<n}a_n ) := \frac{ \rho (x_{1:n} \cbar a_{1:n}) }{ \rho( x_{<n} \cbar a_{<n} ) }\label{cond prob of or}
\end{equation}
$\forall a_{1:n} \forall x_{1:n}$ such that $\rho( x_{<n} \cbar a_{<n} ) > 0$.
It now follows that
\begin{equation}
\label{eq:ChainRule}
\rho(x_{1:n} \cbar a_{1:n}) = \rho( x_1 \cbar a_1) \rho( x_2 \cbar ax_1a_2) \cdots \rho( x_n \cbar ax_{<n}a_n).
\end{equation}

Definition \ref{def:environment_model} is used in two distinct ways. The first is a means of describing the true underlying environment.
This may be unknown to the agent.
Alternatively, we can use Definition \ref{def:environment_model} to describe an agent's \emph{subjective} model of the environment.
This model is typically learnt, and will often only be an approximation to the true environment.
To make the distinction clear, we will refer to an agent's \emph{environment model} when talking about the agent's model of the environment.

Notice that $\rho(\cdot \cbar h)$ can be an arbitrary function of the agent's previous history $h$.
Our definition of environment is sufficiently general to encapsulate a wide variety of environments, including standard reinforcement learning setups such as MDPs or POMDPs.

\paradot{Reward, Policy and Value Functions}\label{sec:RewardValueFunction}
We now cast the familiar notions of \emph{reward}, \emph{policy} and \emph{value} \cite{sutton-barto98} into our setup.
The agent's goal is to accumulate as much reward as it can during its lifetime.
More precisely, the agent seeks a \emph{policy} that will allow it to maximise its expected future reward up to a fixed, finite, but
arbitrarily large horizon $m \in \mathbb{N}$.
The instantaneous reward values are assumed to be bounded.
Formally, a policy is a function that maps a history to an action.
If we define $R_k(aor_{\leq t}) := r_k$ for $1 \leq k \leq t$, then we have the following definition for the expected
future value of an agent acting under a particular policy:

\begin{defn}\label{defn:policy_value}
Given history $ax_{1:t}$, the $m$-horizon expected future reward of an agent acting under policy $\pi \colon (\mathcal{A} \times \mathcal{X})^* \rightarrow \cA$ with respect to an environment $\rho$ is:
\begin{equation}\label{eq:val}
v_{\rho}^m(\pi, ax_{1:t}) := \mathbb{E}_{\rho} \left[ \sum\limits_{i=t+1}^{t+m} R_i(ax_{\leq t+m} ) \biggm\vert x_{1:t} \right],
\end{equation}
where for $t < k \leq t+m$, $a_k := \pi(ax_{<k})$.
The quantity $v_{\rho}^m(\pi, ax_{1:t}a_{t+1})$ is defined similarly, except that $a_{t+1}$ is now no longer defined by $\pi$.
\end{defn}

The optimal policy $\pi^*$ is the policy that maximises the expected future reward.
The maximal achievable expected future reward of an agent with history $h$ in environment $\rho$ looking $m$ steps ahead is $V_{\rho}^m(h) := v_{\rho}^m(\pi^*, h)$.
It is easy to see that if $h \in ({\cal A} \times {\cal X})^{t}$, then
\begin{equation}\label{value defn}
V_{\rho}^{m}(h) =
 \max\limits_{a_{t+1}}\sum\limits_{x_{t+1}}{\rho(x_{t+1} \cbar h a_{t+1})} \cdots
\max\limits_{a_{t+m}}\sum\limits_{x_{t+m}}{\rho(x_{t+m} \cbar h ax_{t+1:t+m-1} a_{t+m})}
\left[\sum\limits_{i=t+1}^{t+m}r_i\right].
\end{equation}

For convenience, we will often refer to Equation (\ref{value defn}) as the \emph{expectimax operation}.
Furthermore, the $m$-horizon optimal action $a^*_{t+1}$ at time $t+1$ is related to the expectimax operation by
\begin{equation}\label{eq:optimal_action_known_env}
a^*_{t+1} = \arg\max\limits_{a_{t+1}}V_{\rho}^{m}(ax_{1:t}a_{t+1}).
\end{equation}

Equations (\ref{eq:val}) and (\ref{value defn}) can be modified to handle discounted reward, however we focus on the finite-horizon case since it both aligns with AIXI and allows for a simplified presentation.

\section{Bayesian Agents}\label{sec:BayesAgents}

As mentioned earlier, Definition \ref{def:environment_model} can be used to describe the agent's subjective model of the true environment.
Since we are assuming that the agent does not initially know the true environment, we desire subjective models whose predictive performance improves as the agent gains experience.
One way to provide such a model is to take a Bayesian perspective.
Instead of committing to any single fixed environment model, the agent uses a \emph{mixture} of environment models.
This requires committing to a class of possible environments (the model class), assigning an initial weight to each possible environment (the prior), and subsequently updating the weight for each model using Bayes rule (computing the posterior) whenever more experience is obtained.
The process of learning is thus implicit within a Bayesian setup.

The mechanics of this procedure are reminiscent of Bayesian methods to predict sequences of (single typed) observations.
The key difference in the agent setup is that each prediction may now also depend on previous agent actions.
We incorporate this by using the \emph{action conditional} definitions and identities of Section \ref{sec:agent setting}.

\begin{defn}
\label{def:mixture_environment}
Given a countable model class $\mathcal{M} := \{ \rho_1, \rho_2, \dots\}$ and a prior weight $w_0^{\rho}>0$ for each $\rho \in \mathcal{M}$ such that $\sum_{\rho\in\mathcal{M}}w_0^\rho=1$,
the mixture environment model is  $\xi(x_{1:n} \cbar a_{1:n}) := \sum\limits_{\rho \in \mathcal{M}} w_0^{\rho} \rho(x_{1:n} \cbar a_{1:n})$.
\end{defn}

The next proposition allows us to use a mixture environment model whenever we can use an environment model.

\begin{proposition}
\label{prop:MixtureEnvModelsAreEnvModels}
A mixture environment model is an environment model.
\begin{proof}
$\forall a_{1:n} \in \cA^n$ and $\forall x_{<n} \in \cX^{n-1}$ we have that
\begin{eqnarray*}
\sum\limits_{x_n \in \cX} \xi(x_{1:n} \cbar a_{1:n}) =
\sum\limits_{x_n \in \cX} \sum\limits_{\rho \in \mathcal{M}} w_0^{\rho} \rho(x_{1:n} \cbar a_{1:n}) =
\sum\limits_{\rho \in \mathcal{M}} w_0^{\rho} \sum\limits_{x_n \in \cX} \rho(x_{1:n} \cbar a_{1:n}) =
\xi(x_{<n} \cbar a_{<n})
\end{eqnarray*}
where the final step follows from application of Equation (\ref{eq:chronological}) and Definition \ref{def:mixture_environment}.
\end{proof}
\end{proposition}

The importance of Proposition \ref{prop:MixtureEnvModelsAreEnvModels} will become clear in the context of planning with environment models, described in Section \ref{sec:mcts}.

\paradot{Prediction with a Mixture Environment Model}
As a mixture environment model is an environment model, we can simply use:
\begin{equation}
\label{eq:MixturePredictor}
\xi(x_n \cbar ax_{<n}a_n) = \frac{\xi(x_{1:n} \cbar a_{1:n})}{\xi(x_{<n} \cbar a_{<n})}
\end{equation}
to predict the next observation reward pair.
Equation (\ref{eq:MixturePredictor}) can also be expressed in terms of a convex combination of model predictions, with each model weighted by its posterior, from
\begin{gather*}
\label{eq:PosteriorPredictor}
\xi(x_n \cbar ax_{<n}a_n) =
\frac{\sum\limits_{\rho \in \mathcal{M}} w_0^{\rho} \rho(x_{1:n} \cbar a_{1:n})}{\sum\limits_{\rho \in \mathcal{M}} w_0^{\rho} \rho(x_{<n} \cbar a_{<n})}
= \sum\limits_{\rho \in \mathcal{M}} w_{n-1}^{\rho} \rho(x_n \cbar ax_{<n}a_n),
\end{gather*}
where the posterior weight $w_{n-1}^{\rho}$ for environment model $\rho$ is given by
\begin{equation}
\label{eq:PosteriorWeight}
w_{n-1}^{\rho} := \frac{w_0^{\rho} \rho(x_{<n} \cbar a_{<n})}{\sum\limits_{\nu \in \mathcal{M}} w_0^{\nu} \nu(x_{<n} \cbar a_{<n})} = \Pr(\rho \cbar ax_{<n})
\end{equation}

If $|\mathcal{M}|$ is finite, Equations (\ref{eq:MixturePredictor}) and (\ref{eq:PosteriorPredictor}) can be maintained online in $O(|\mathcal{M}|)$ time by using the fact that
\begin{equation*}
\rho(x_{1:n} \cbar a_{1:n}) = \rho(x_{<n} \cbar a_{<n}) \rho(x_n \cbar ax_{<n}a),
\end{equation*}
which follows from Equation (\ref{eq:ChainRule}), to incrementally maintain the likelihood term for each model.

\paradot{Theoretical Properties}
We now show that if there is a good model of the (unknown) environment in $\mathcal{M}$, an agent using the mixture environment model
\begin{equation}
\xi(x_{1:n} \cbar a_{1:n}) := \sum\limits_{\rho \in \mathcal{M}} w_0^{\rho} \rho(x_{1:n} \cbar a_{1:n})
\end{equation}
will predict well.
Our proof is an adaptation from \citeA{Hutter:04uaibook}.
We present the full proof here as it is both instructive and directly relevant to many different kinds of practical Bayesian agents.

First we state a useful entropy inequality.

\begin{lem}[\citeA{Hutter:04uaibook}]\label{entropy ineq}
Let $\{y_i\}$ and $\{ z_i \}$ be two probability distributions, i.e.\ $y_i \geq 0, z_i \geq 0,$ and $\sum_i y_i = \sum_i z_i = 1$.
Then we have
\begin{equation*}
\sum_{i} (y_i - z_i)^2 \leq \sum_i y_i \ln \frac{y_i}{z_i}.
\end{equation*}
\end{lem}

\begin{thm}\label{conv to mu}
Let $\mu$ be the true environment.
The $\mu$-expected squared difference of $\mu$ and $\xi$ is bounded as follows.
For all $n \in \mathbb{N}$, for all $a_{1:n}$,
\begin{gather*}
\sum_{k=1}^n \sum_{x_{1:k}} \mu(x_{<k} \cdbar a_{<k}) \biggl( \mu(x_k \cbar ax_{<k}a_k) - \xi( x_k \cbar ax_{<k}a_k ) \biggr)^2
 \leq \min_{\rho \in \mathcal{M}} \,\biggl\{\, -\ln w_0^{\rho} + D_{1:n}(\mu \,\|\, \rho) \,\biggr\},
\end{gather*}
where $D_{1:n}(\mu \,\|\, \rho) := \sum_{x_{1:n}} \mu( x_{1:n} \cdbar a_{1:n}) \ln \frac{\mu( x_{1:n} \cdbar a_{1:n})} {\rho( x_{1:n} \cdbar a_{1:n})}$ is the KL divergence of  $\mu(\cdot \cdbar a_{1:n})$ and $\rho( \cdot \cdbar a_{1:n})$.
\end{thm}
{\allowdisplaybreaks
\begin{proof}
Combining \citeA[\S 3.2.8 and \S 5.1.3]{Hutter:04uaibook} we get
\begin{align*}
&\;\;\;\; \sum_{k=1}^n \sum_{x_{1:k}} \mu(x_{<k} \cdbar a_{<k}) \biggl( \mu(x_k \cbar ax_{<k}a_k) - \xi( x_k \cbar ax_{<k}a_k ) \biggr)^2 \\
 &= \sum_{k=1}^n \sum_{x_{<k}} \mu(x_{<k} \cdbar a_{<k}) \sum_{x_k} \biggl( \mu(x_k \cbar ax_{<k}a_k) - \xi( x_k \cbar ax_{<k}a_k ) \biggr)^2 \\
 &\leq \sum_{k=1}^n \sum_{x_{<k}} \mu(x _{<k} \cdbar a_{<k}) \sum_{x_k} \mu(x_k \cbar ax_{<k}a_k) \ln \frac{ \mu (x_k \cbar ax_{<k}a_k) }{ \xi( x_k \cbar
   ax_{<k}a_k ) } \hspace{2em} \text{[Lemma \ref{entropy ineq}]} \\
 &= \sum_{k=1}^n \sum_{x_{1:k}} \mu( x_{1:k} \cdbar a_{1:k}) \ln \frac{ \mu( x_k \cbar ax_{<k}a_k) }{ \xi( x_k \cbar ax_{<k}a_k ) } \hspace{8.8em} \text{[Equation (\ref{cond prob of or})]}\\
 &= \sum_{k=1}^n \sum_{x_{1:k}} \biggl( \sum_{x_{k+1:n}} \mu( x_{1:n} \cdbar a_{1:n}) \biggr) \ln \frac{ \mu( x_k \cbar ax_{<k}a_k) }{
   \xi( x_k \cbar ax_{<k}a_k) } \hspace{6.3em} \text{[Equation (\ref{eq:chronological})]} \\
 &= \sum_{k=1}^n \sum_{x_{1:n}} \mu( x_{1:n} \cdbar a_{1:n}) \ln \frac{ \mu( x_k \cbar ax_{<k}a_k) }{ \xi( x_k \cbar ax_{<k}a_k) } \\
 &= \sum_{x_{1:n}} \mu( x_{1:n} \cdbar a_{1:n}) \sum_{k=1}^n \ln \frac{ \mu( x_k \cbar ax_{<k}a_k) }{ \xi( x_k \cbar ax_{<k}a_k) } \\
 &= \sum_{x_{1:n}} \mu( x_{1:n} \cdbar a_{1:n}) \ln \frac{ \mu( x_{1:n} \cdbar a_{1:n}) }{ \xi( x_{1:n} \cdbar a_{1:n}) } \hspace{10.95em} \text{[Equation (\ref{eq:ChainRule})]}\\
 &= \sum_{x_{1:n}} \mu( x_{1:n} \cdbar a_{1:n}) \ln \left[ \frac{\mu(x_{1:n}\cdbar a_{1:n})}{\rho( x_{1:n} \cdbar a_{1:n})}  \frac{\rho( x_{1:n} \cdbar
    a_{1:n})} {\xi( x_{1:n} \cdbar a_{1:n})} \right]  \hspace{5.1em} \text{[arbitrary $\rho \in \mathcal{M}$]}\\
 &= \sum_{x_{1:n}} \mu( x_{1:n} \cdbar a_{1:n}) \ln \frac{\mu(x_{1:n}\cdbar a_{1:n})}{\rho( x_{1:n} \cdbar a_{1:n})} +
    \sum_{x_{1:n}} \mu( x_{1:n} \cdbar a_{1:n}) \ln \frac{\rho( x_{1:n} \cdbar
     a_{1:n})} {\xi( x_{1:n} \cdbar a_{1:n})}\\
 &\leq D_{1:n}( \mu \,\|\, \rho) + \sum_{x_{1:n}} \mu( x_{1:n} \cdbar a_{1:n}) \ln \frac{ \rho( x_{1:n} \cdbar a_{1:n}) }{ w_0^{\rho} \rho( x_{1:n} \cdbar a_{1:n}) } \hspace{4.5em} \text{[Definition \ref{def:mixture_environment}]} \\
 &= D_{1:n}( \mu \,\|\, \rho) - \ln w_0^{\rho}.
\end{align*}
Since the inequality holds for arbitrary $\rho \in \mathcal{M}$, it holds for the minimising $\rho$.
\end{proof}

In Theorem~\ref{conv to mu}, take the supremum over $n$ in the r.h.s and then the limit $n \rightarrow \infty$ on the l.h.s.
If $\sup_n D_{1:n}( \mu \,\|\, \rho) < \infty$ for the minimising $\rho$, the infinite sum on the l.h.s can only be finite if $\xi( x_k \cbar ax_{<k}a_k)$ converges sufficiently fast to $\mu( x_k \cbar ax_{<k}a_k)$ for $k \rightarrow \infty$ with probability 1, hence $\xi$ predicts $\mu$ with rapid convergence.
As long as $D_{1:n}( \mu \,\|\, \rho) = o(n)$, $\xi$ still converges to $\mu$ but in a weaker Ces{\`a}ro sense.
The contrapositive of the statement tells us that if $\xi$ fails to predict the environment well, then there is no good model in $\mathcal{M}$.

\paradot{AIXI: The Universal Bayesian Agent}
Theorem \ref{conv to mu} motivates the construction of Bayesian agents that use rich model classes.
The AIXI agent can be seen as the limiting case of this viewpoint, by using the largest model class expressible on a Turing machine.

Note that AIXI can handle stochastic environments since Equation~(\ref{aixi_eq}) can be shown to be formally equivalent to
\begin{equation}
\label{aixi_eq2}
a_t^* = \arg\max\limits_{a_t}\sum\limits_{o_tr_t} \dots \max\limits_{a_{t+m}} \sum\limits_{o_{t+m}r_{t+m}}
[r_t + \dots + r_{t+m}]
\sum\limits_{\rho \in {{\cal M}_U}}2^{-K(\rho)} \rho( x_{1:t+m} \cdbar a_{1:t+m}),
\end{equation}
where $\rho( x_{1:t+m} \cdbar a_1\ldots a_{t+m})$ is the probability of observing $x_1x_2 \ldots x_{t+m}$ given actions $a_1a_2\ldots a_{t+m}$,
class ${\cal M}_U$ consists of all enumerable chronological semimeasures \cite{Hutter:04uaibook}, which includes all computable $\rho$, and $K(\rho)$ denotes the Kolmogorov complexity \cite{li-vitanyi} of $\rho$ with respect to $U$.
In the case where the environment is a computable function and
\begin{equation}
\xi_U(x_{1:t} \cdbar a_{1:t}) := \sum\limits_{\rho \in {\cal M}_U}2^{-K(\rho)} \rho( x_{1:t} \cdbar a_{1:t}),
\end{equation}
Theorem~\ref{conv to mu} shows for all $n \in \mathbb{N}$ and for all $a_{1:n}$,
\begin{gather}
\label{eq:AIXI_bound}
\sum_{k=1}^n \sum_{x_{1:k}} \mu(x_{<k} \cdbar a_{<k}) \biggl( \mu(x_k \cbar ax_{<k}a_k) - \xi_U( x_k \cbar ax_{<k}a_k ) \biggr)^2
\leq K(\mu) \ln 2.
\end{gather}

\paradot{Direct AIXI Approximation}
We are now in a position to describe our approach to AIXI approximation.
For prediction, we seek a computationally efficient mixture environment model $\xi$ as a replacement for $\xi_U$.
Ideally, $\xi$ will retain $\xi_U$'s bias towards simplicity and some of its generality.
This will be achieved by placing a suitable Ockham prior over a set of candidate environment models.

For planning, we seek a scalable algorithm that can, given a limited set of resources, compute an approximation to the expectimax action given by
\begin{equation*}
a^*_{t+1} = \arg\max\limits_{a_{t+1}}V_{\xi_U}^{m}(ax_{1:t}a_{t+1}).
\end{equation*}

The main difficulties are of course computational.
The next two sections introduce two algorithms that can be used to (partially) fulfill these criteria.
Their subsequent combination will constitute our AIXI approximation.

\section{Expectimax Approximation with Monte-Carlo Tree Search}\label{sec:mcts}

Na{\" i}ve computation of the expectimax operation (Equation \ref{value defn}) takes $O(|\cA \times \cX|^m)$ time, unacceptable for all but tiny values of $m$.
This section introduces \searchalg, a generalisation of the popular Monte-Carlo Tree Search algorithm UCT \cite{kocsis06}, that can be used to approximate a finite horizon expectimax operation given an environment model $\rho$.
As an environment model subsumes both MDPs and POMDPs, \searchalg\ effectively extends the UCT algorithm to a wider class of problem domains.

\paradot{Background}
UCT has proven particularly effective in dealing with difficult problems containing large state spaces.
It requires a generative model that when given a state-action pair $(s, a)$ produces a subsequent state-reward pair $(s', r)$ distributed according to $\Pr(s',r \cbar s, a)$.
By successively sampling trajectories through the state space, the UCT algorithm incrementally constructs a search tree, with each node containing an estimate of the value of each state.
Given enough time, these estimates converge to their true values.

The \searchalg\ algorithm can be realised by replacing the notion of state in UCT by an agent history $h$ (which is always a sufficient statistic) and using an environment model $\rho$ to predict the next percept.
The main subtlety with this extension is that now the history condition of the percept probability $\rho(or \cbar h)$ needs to be updated during the search.
This is to reflect the extra information an agent will have at a hypothetical future point in time.
Furthermore, ~Proposition~\ref{prop:MixtureEnvModelsAreEnvModels} allows \searchalg\ to be instantiated with a mixture environment model, which directly incorporates the model uncertainty of the agent into the planning process.
This gives (in principle, provided that the model class contains the true environment and ignoring issues of limited computation) the well known Bayesian solution to the exploration/exploitation dilemma; namely, if a reduction in model uncertainty would lead to higher expected future reward, \searchalg\ would recommend an information gathering action.

\paradot{Overview}
\searchalg\  is a best-first Monte-Carlo Tree Search technique that iteratively constructs a search
tree in memory.
The tree is composed of two interleaved types of nodes: decision nodes and chance nodes.
These correspond to the alternating max and sum operations in the expectimax operation.
Each node in the tree corresponds to a history $h$.
If $h$ ends with an action, it is a chance node; if $h$ ends with an observation-reward pair, it is a decision node.
Each node contains a statistical estimate of the future reward.

Initially, the tree starts with a single decision node containing $|\cA|$ children.
Much like existing MCTS methods \cite{chaslot08d}, there are four conceptual phases to a single iteration of \searchalg.
The first is the {\em selection} phase, where the search tree is traversed from the root node to an existing leaf chance node $n$.
The second is the {\em expansion} phase, where a new decision node is added as a child to $n$.
The third is the {\em simulation} phase, where a rollout policy in conjunction with the environment model $\rho$ is used to sample a possible future path from $n$ until a fixed distance from the root is reached. Finally, the {\em backpropagation} phase updates the value estimates for each node on the reverse trajectory leading back to the root.
Whilst time remains, these four conceptual operations are repeated. Once the time limit is reached, an approximate best action can be selected by looking at the value estimates of the children of the root node.

During the selection phase, action selection at decision nodes is done using a policy that balances exploration and exploitation.
This policy has two main effects:
\begin{itemize}\itemsep1mm\parskip0mm
  \item to gradually move the estimates of the future reward towards the maximum attainable future reward if the agent acted optimally.
  \item to cause asymmetric growth of the search tree towards areas that have high predicted reward, implicitly pruning large parts of the search space.
\end{itemize}

The future reward at leaf nodes is estimated by choosing actions according to a heuristic policy until a total of $m$ actions have been made by the agent, where $m$ is the search horizon. This heuristic estimate helps the agent to focus its exploration on useful parts of the search tree, and in practice allows for a much larger horizon than a brute-force expectimax search.

\searchalg\  builds a sparse search tree in the sense that observations are only added to chance nodes once they have been generated along some sample path.
A full-width expectimax search tree would not be sparse; each possible stochastic outcome would be represented by a distinct node in the search tree.
For expectimax, the branching factor at chance nodes is thus $|O|$, which means that searching to even moderate sized $m$ is intractable.

Figure \ref{staructtree} shows an example \searchalg\  tree.
Chance nodes are denoted with stars.
Decision nodes are denoted by circles.
The dashed lines from a star node indicate that not all of the children have been expanded. The squiggly line at the base of the leftmost leaf denotes
the execution of a rollout policy.
The arrows proceeding up from this node indicate the flow of information back up the tree; this is defined in more detail below.

\begin{figure*}
\centerline{\mbox{\includegraphics[scale=1.4]{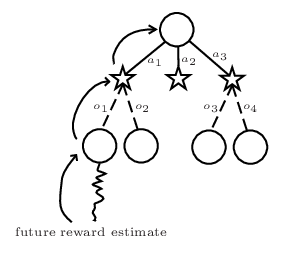}}}
\vspace{-1em}
\caption{A \searchalg\ search tree}
\label{staructtree}
\vspace{-0.5em}
\end{figure*}

\paradot{Action Selection at Decision Nodes}
A decision node will always contain $|\mathcal{A}|$ distinct children, all of whom are chance nodes.
Associated with each decision node representing a particular history $h$ will be a value function estimate, $\hat{V}(h)$.
During the selection phase, a child will need to be picked for further exploration.
Action selection in MCTS poses a classic exploration/exploitation dilemma.
On one hand we need to allocate enough visits to all children to ensure that we have accurate estimates for them, but on the other hand we need to allocate enough visits to the maximal action to ensure convergence of the node to the value of the maximal child node.

Like UCT, \searchalg\  recursively uses the UCB policy \cite{auer02} from the $n$-armed bandit setting at each decision node to determine
which action needs further exploration. Although the uniform logarithmic regret bound no longer carries across from the bandit setting, the UCB policy has been
shown to work well in practice in complex domains such as computer Go \cite{Gelly06} and General Game Playing \cite{cadia2008}. This policy has the advantage of
ensuring that at each decision node, every action eventually gets explored an infinite number of times, with the best action being selected exponentially more
often than actions of lesser utility.

\begin{defn}
The visit count $T(h)$ of a decision node $h$ is the number of times $h$ has been sampled by the \searchalg\  algorithm.
The visit count of the chance node found by taking action $a$ at $h$ is defined similarly, and is denoted by $T(ha)$.
\end{defn}

\begin{defn}
\label{eq:ucb}
Suppose $m$ is the remaining search horizon and each instantaneous reward is bounded in the interval $[\alpha, \beta]$.
Given a node representing a history $h$ in the search tree, the action picked by the UCB action selection policy is:
\begin{equation}
a_{UCB}(h) :=
\arg\max_{a \in \mathcal{A}}
\begin{cases}
  \frac{1}{m(\beta - \alpha)} \hat{V}(ha) + C \sqrt{\frac{\log(T(h))}{T(ha)}} & \mbox{if } T(ha) > 0;\\
  \infty & \mbox{otherwise},
\end{cases}
\end{equation}
where $C \in \mathbb{R}$ is a positive parameter that controls the ratio of exploration to exploitation.
If there are multiple maximal actions, one is chosen uniformly at random.
\end{defn}
Note that we need a linear scaling of $\hat{V}(ha)$ in Definition~\ref{eq:ucb} because the UCB policy is only applicable for rewards confined to the $[0,1]$ interval.

\paradot{Chance Nodes}
Chance nodes follow immediately after an action is selected from a decision node.
Each chance node $ha$ following a decision node $h$ contains an estimate of the future utility denoted by $\hat{V}(ha)$.
Also associated with the chance node $ha$  is a density $\rho(\cdot \cbar ha)$ over observation-reward pairs.

After an action $a$ is performed at node $h$, $\rho(\cdot \cbar ha)$ is sampled once to generate the next observation-reward pair $or$.
If $or$ has not been seen before, the node $haor$ is added as a child of $ha$.

\paradot{Estimating Future Reward at Leaf Nodes}\label{subsec:playout}
If a leaf decision node is encountered at depth $k < m$ in the tree, a means of estimating the future reward for the remaining $m-k$ time steps is required.
MCTS methods use a heuristic rollout policy $\Pi$ to estimate the sum of future rewards $\sum_{i=k}^m r_i$.
This involves sampling an action $a$ from $\Pi(h)$, sampling a percept $or$ from $\rho(\cdot \cbar ha)$, appending $aor$ to the current history $h$ and then repeating this process until the horizon is reached.
This procedure is described in Algorithm \ref{alg:playout}.
A natural baseline policy is $\Pi_{random}$, which chooses an action uniformly at random at each time step.

As the number of simulations tends to infinity, the structure of the \searchalg\ search tree converges to the full depth $m$ expectimax tree.
Once this occurs, the rollout policy is no longer used by \searchalg.
This implies that the asymptotic value function estimates of \searchalg\ are invariant to the choice of $\Pi$.
In practice, when time is limited, not enough simulations will be performed to grow the full expectimax tree.
Therefore, the choice of rollout policy plays an important role in determining the overall performance of \searchalg.
Methods for learning $\Pi$ online are discussed as future work in Section \ref{sec:future_work}.
Unless otherwise stated, all of our subsequent results will use $\Pi_{random}$.

\paradot{Reward Backup}\label{backup}
After the selection phase is completed, a path of nodes $n_1 n_2 \dots n_k$, $k \leq m$, will have been traversed from the root of the search
tree $n_1$ to some leaf $n_k$.
For each $1 \leq j \leq k$, the statistics maintained for history $h_{n_j}$ associated with node $n_j$ will be updated as follows:

\begin{equation}\label{eq:backup}
\hat{V}(h_{n_j}) \leftarrow \frac{T(h_{n_j})}{T(h_{n_j})+1}\hat{V}(h_{n_j})+\frac{1}{T(h_{n_j})+1} \sum\limits_{i=j}^{m} r_i
\end{equation}
\begin{equation}\label{eq:inc}
T(h_{n_j}) \leftarrow T(h_{n_j}) + 1
\end{equation}
Equation (\ref{eq:backup}) computes the mean return.
Equation (\ref{eq:inc}) increments the visit counter.
Note that the same backup operation is applied to both decision and chance nodes.

\paradot{Pseudocode}
The pseudocode of the \searchalg\ algorithm is now given.

After a percept has been received, Algorithm \ref{alg:rhoUCT} is invoked to determine an approximate best action.
A \emph{simulation} corresponds to a single call to {\sc Sample} from Algorithm \ref{alg:rhoUCT}.
By performing a number of simulations, a search tree $\Psi$ whose root corresponds to the current history $h$ is constructed.
This tree will contain estimates $\hat{V}_{\rho}^m(ha)$ for each $a\in \mathcal{A}$.
Once the available thinking time is exceeded, a maximising action $\hat{a}_h^* := \arg\max_{a \in \cA} \hat{V}_{\rho}^m(ha)$ is retrieved by {\sc BestAction}.
Importantly, Algorithm \ref{alg:rhoUCT} is \emph{anytime}, meaning that an approximate best action is always available.
This allows the agent to effectively utilise all available computational resources for each decision.

\algsetup{indent=2em}
\begin{algorithm}[h!]
\caption{\searchalg$(h, m)$}\label{alg:rhoUCT}
\begin{algorithmic}[1]
\REQUIRE A history $h$
\REQUIRE A search horizon $m \in \mathbb{N}$
\medskip
\STATE {\sc Initialise}$(\Psi)$
\REPEAT
  \STATE {\sc Sample}$(\Psi, h, m)$
\UNTIL {out of time}
\RETURN {\sc BestAction}$(\Psi,h)$
\end{algorithmic}
\end{algorithm}

For simplicity of exposition, {\sc Initialise} can be understood to simply clear the entire search tree $\Psi$.
In practice, it is possible to carry across information from one time step to another.
If $\Psi_t$ is the search tree obtained at the end of time $t$, and $aor$ is the agent's actual action and experience at time $t$,
then we can keep the subtree rooted at node $\Psi_t(hao)$ in $\Psi_t$ and make that the search tree $\Psi_{t+1}$ for use at the beginning of the next time step.
The remainder of the nodes in $\Psi_t$ can then be deleted.

Algorithm \ref{alg:sample} describes the recursive routine used to sample a single future trajectory.
It uses the {\sc SelectAction} routine to choose moves at decision nodes, and invokes the {\sc Rollout} routine at unexplored leaf nodes.
The {\sc Rollout} routine picks actions according to the rollout policy $\Pi$ until the (remaining) horizon is reached, returning the accumulated reward.
After a complete trajectory of length $m$ is simulated, the value estimates are updated for each node traversed as per Section \ref{backup}.
Notice that the recursive calls on Lines \ref{alg:sample:rec1} and \ref{alg:sample:rec2} append the most recent percept or action to the history argument.

\algsetup{indent=2em}
\begin{algorithm}[h!]
\caption{{\sc Sample}$(\Psi, h, m)$}\label{alg:sample}
\begin{algorithmic}[1]
\REQUIRE A search tree $\Psi$
\REQUIRE A history $h$
\REQUIRE A remaining search horizon $m \in \mathbb{N}$
\medskip
\IF {$m=0$}
  \RETURN 0
\ELSIF {$\Psi(h)$ is a chance node}
  \STATE  Generate $(o,r)$ from $\rho(or \cbar h)$
  \STATE  Create node $\Psi(hor)$ if $T(hor) = 0$
  \STATE  reward $\leftarrow$ $r$ $+$ {\sc Sample}$(\Psi, hor, m - 1)$\label{alg:sample:rec1}
\ELSIF {$T(h)=0$}
  \STATE reward $\leftarrow$ {\sc Rollout}$(h, m)$
\ELSE
  \STATE $a$ $\leftarrow$ {\sc SelectAction}$(\Psi, h)$
  \STATE reward $\leftarrow$ {\sc Sample}$(\Psi, ha, m)$\label{alg:sample:rec2}
\ENDIF
\STATE $\hat{V}(h) \leftarrow \frac{1}{T(h)+1}[reward + T(h)\hat{V}(h)]$
\STATE $T(h) \leftarrow T(h)+1$
\RETURN reward
\end{algorithmic}
\end{algorithm}

The action chosen by {\sc SelectAction} is specified by the UCB policy described in Definition \ref{eq:ucb}.
If the selected child has not been explored before, a new node is added to the search tree.
The constant $C$ is a parameter that is used to control the shape of the search tree; lower values of $C$ create deep, selective search trees, whilst higher values lead to shorter, bushier trees.
UCB automatically focuses attention on the best looking action in such a way that the sample estimate $\hat{V}_\rho(h)$ converges to $V_\rho(h)$, whilst still exploring alternate actions sufficiently often to guarantee that the best action will be eventually found.

\algsetup{indent=2em}
\begin{algorithm}[h!]
\caption{{\sc SelectAction}$(\Psi, h)$}\label{alg:select_action}
\begin{algorithmic}[1]
\REQUIRE A search tree $\Psi$
\REQUIRE A history $h$
\REQUIRE An exploration/exploitation constant $C$
\medskip
\STATE $\mathcal{U} = \{ a\in \cA \colon T(ha) = 0\}$
\IF {$\mathcal{U} \ne \{\}$}
  \STATE Pick $a \in \mathcal{U}$ uniformly at random
  \STATE Create node $\Psi(ha)$
  \RETURN a
\ELSE
  \RETURN $\arg\max\limits_{a \in \cA} \left\{ \frac{1}{m(\beta - \alpha)} \hat{V}(ha) + C \sqrt{\frac{\log(T(h))}{T(ha)}} \right\}$
\ENDIF
\end{algorithmic}
\end{algorithm}

\algsetup{indent=2em}
\begin{algorithm}[h!]
\caption{{\sc Rollout}$(h, m)$}\label{alg:playout}
\begin{algorithmic}[1]
\REQUIRE A history $h$
\REQUIRE A remaining search horizon $m \in \mathbb{N}$
\REQUIRE A rollout function $\Pi$
\medskip
\STATE $reward \leftarrow 0$
\FOR {$i=1$ to $m$}
  \STATE Generate $a$ from $\Pi(h)$
  \STATE Generate $(o,r)$ from $\rho(or \cbar ha)$
  \STATE $reward \leftarrow reward + r$
  \STATE $h \leftarrow haor$
\ENDFOR
\RETURN reward
\end{algorithmic}
\end{algorithm}

\paradot{Consistency of \searchalg}
Let $\mu$ be the true underlying environment.
We now establish the link between the expectimax value $V_{\mu}^{m}(h)$ and its estimate $\hat{V}_\mu^m(h)$ computed by the \searchalg\ algorithm.

\citeA{kocsis06} show that with an appropriate choice of $C$, the UCT algorithm is consistent in finite horizon MDPs.
By interpreting histories as Markov states, our general agent problem reduces to a finite horizon MDP.
This means that the results of \citeA{kocsis06} are now directly applicable.
Restating the main consistency result in our notation, we have
\begin{equation}\label{uctc}
\forall \epsilon \forall h \lim_{T(h) \to \infty} \Pr \left( | V_{\mu}^m(h) - \hat{V}_\mu^m(h) | \leq \epsilon \right) = 1,
\end{equation}
that is, $\hat{V}_\mu^m(h) \to V_{\mu}^m(h) $ with probability $1$.
Furthermore, the probability that a suboptimal action (with respect to $V^m_\mu(\cdot)$) is picked by \searchalg\ goes to zero in the limit.
Details of this analysis can be found in \cite{kocsis06}.

\paradot{Parallel Implementation of \searchalg}
As a Monte-Carlo Tree Search routine, Algorithm \ref{alg:rhoUCT} can be easily parallelised.
The main idea is to concurrently invoke the {\sc Sample} routine whilst providing appropriate locking mechanisms for the interior nodes of the search tree.
A highly scalable parallel implementation is beyond the scope of the paper, but it is worth noting that ideas applicable to
high performance Monte-Carlo Go programs \cite{chaslot08} can be easily transferred to our setting.

\section{Model Class Approximation using Context Tree Weighting}\label{sec:ctw}

We now turn our attention to the construction of an efficient mixture environment model suitable for the general reinforcement learning problem.
If computation were not an issue, it would be sufficient to first specify a large model class $\mathcal{M}$, and then use Equations (\ref{eq:MixturePredictor}) or (\ref{eq:PosteriorPredictor}) for online prediction.
The problem with this approach is that at least $O(|\mathcal{M}|)$ time is required to process each new piece of experience.
This is simply too slow for the enormous model classes required by general agents.
Instead, this section will describe how to predict in $O(\log\log|\mathcal{M}|)$ time, using a mixture environment model constructed from an adaptation of the Context Tree Weighting algorithm.

\paradot{Context Tree Weighting}
Context Tree Weighting (CTW) \cite{ctw95,ctw-tutorial} is an efficient and theoretically well-studied binary sequence prediction
algorithm that works well in practice \cite{begleiter04}.
It is an online Bayesian model averaging algorithm that computes, at each time point $t$, the probability
\begin{equation}\label{eq:ctw pr} \Pr (y_{1:t}) = \sum_{M} \Pr(M) \Pr( y_{1:t} \cbar M), \end{equation}
where $y_{1:t}$ is the binary sequence seen so far, $M$ is a prediction suffix tree \cite{rissanen83,ron96},
$\Pr(M)$ is the prior probability of $M$, and the summation is over \emph{all} prediction suffix trees of bounded depth $D$.
This is a huge class, covering all $D$-order Markov processes.
A na{\" i}ve computation of (\ref{eq:ctw pr}) takes time $O(2^{2^D})$; using CTW, this computation requires only $O(D)$ time.
In this section, we outline two ways in which CTW can be generalised to compute probabilities of the form
\begin{equation}\label{eq:gen ctw pr} \Pr ( x_{1:t} \cbar a_{1:t}) = \sum_{M} \Pr(M) \Pr( x_{1:t} \cbar M, a_{1:t}), \end{equation}
where $x_{1:t}$ is a percept sequence, $a_{1:t}$ is an action sequence, and $M$ is a prediction suffix tree
as in (\ref{eq:ctw pr}).
These generalisations will allow CTW to be used as a mixture environment model.

\paradot{Krichevsky-Trofimov Estimator}
We start with a brief review of the KT estimator \cite{kt-estimator} for Bernoulli distributions.
Given a binary string $y_{1:t}$ with $a$ zeros and $b$ ones, the KT estimate of the probability of the next symbol is as follows:
\begin{gather}
\text{Pr}_{kt}( Y_{t+1} = 1 \cbar y_{1:t} ) := \frac{b + 1/2}{a + b + 1} \label{kt update}\\
\text{Pr}_{kt}( Y_{t+1} = 0 \cbar y_{1:t} ) := 1 - \text{Pr}_{kt} ( Y_{t+1} = 1 \cbar y_{1:t}).\label{kt update 0}
\end{gather}
The KT estimator is obtained via a Bayesian analysis by putting an uninformative (Jeffreys Beta(1/2,1/2)) prior $\text{Pr}(\theta)\propto\theta^{-1/2}(1-\theta)^{-1/2}$ on the parameter $\theta\in[0,1]$ of the Bernoulli distribution.
From (\ref{kt update})-(\ref{kt update 0}), we obtain the following expression for the block probability of a string:
\begin{align*}
  \text{Pr}_{kt}( y_{1:t} ) &= \text{Pr}_{kt}(y_1 \cbar \epsilon) \text{Pr}_{kt}(y_2 \cbar y_1) \cdots \text{Pr}_{kt} (y_t \cbar y_{<t}) \\
  &= \textstyle{\int} \theta^b(1-\theta)^a\,\text{Pr}(\theta)\,d\theta.
\end{align*}
Since $\Pr_{kt}(s)$ depends only on the number of zeros $a_s$ and ones $b_s$ in a string $s$, if we let $0^{a}1^{b}$ denote a string with $a$ zeroes
and $b$ ones, then we have
\begin{equation}\label{kt block}
\text{Pr}_{kt}(s) = \text{Pr}_{kt}(0^{a_s}1^{b_s}) = \frac{1/2 (1 + 1/2) \cdots (a_s -  1/2) 1/2 (1 + 1/2) \cdots (b_s - 1/2)} {(a_s + b_s)!}.
\end{equation}
We write $\Pr_{kt}(a,b)$ to denote $\Pr_{kt}(0^a1^b)$ in the following.
The quantity $\Pr_{kt}(a,b)$ can be updated incrementally \cite {ctw95} as follows:
\begin{align}
\text{Pr}_{kt}(a+1,b) &= \frac{a + 1/2}{a + b + 1} \text{Pr}_{kt}(a,b) \\
\text{Pr}_{kt}(a,b+1) &= \frac{b + 1/2}{a + b + 1} \text{Pr}_{kt}(a,b), \label{kt block inc}
\end{align}
with the base case being $\Pr_{kt}(0,0) = 1$.

\paradot{Prediction Suffix Trees}
We next describe prediction suffix trees, which are a form of variable-order Markov models.

In the following, we work with binary trees where all the left edges are labeled 1 and all the right edges are labeled 0.
Each node in such a binary tree $M$ can be identified by a string in $\{0,1\}^*$ as follows:
$\epsilon$ represents the root node of $M$; and if $n \in \{0,1\}^*$ is a node in $M$, then $n1$ and $n0$ represent the
left and right child of node $n$ respectively.
The set of $M$'s leaf nodes $L(M) \subset \{0,1\}^*$ form a complete prefix-free set of strings.
Given a binary string $y_{1:t}$ such that $t \geq$ the depth of $M$,
we define $M(y_{1:t}) := y_ty_{t-1}\ldots y_{t'}$, where $t' \leq t$ is the (unique) positive integer such that
$y_{t}y_{t-1}\ldots y_{t'} \in L(M)$.
In other words, $M(y_{1:t})$ represents the suffix of $y_{1:t}$ that occurs in tree $M$.

\begin{defn}\label{defn:pst}
A prediction suffix tree (PST) is a pair $(M,\Theta)$, where $M$ is a binary tree and
associated with each leaf node $l$ in $M$ is a probability distribution over $\{0,1\}$ parametrised by $\theta_l \in \Theta$.
We call $M$ the model of the PST and $\Theta$ the parameter of the PST, in accordance with the terminology of \citeA{ctw95}.
\end{defn}

A prediction suffix tree $(M,\Theta)$ maps each binary string $y_{1:t}$, where $t \geq$ the depth of $M$, to the
probability distribution $\theta_{M(y_{1:t})}$;
the intended meaning is that $\theta_{M(y_{1:t})}$ is the probability that the next bit following $y_{1:t}$ is 1.
For example, the PST shown in Figure~\ref{fig:pst} maps the string $1110$ to $\theta_{M(1110)} = \theta_{01} = 0.3$, which means the next bit
after $1110$ is $1$ with probability $0.3$.

\begin{figure}
\hspace{2em}\centerline{
\xymatrix @ur {
\theta_1 = 0.1  & {\circ} \ar[l]_1 \ar[d]^0 \\
\theta_{01} = 0.3 & {\circ} \ar[l]_1 \ar[d]^0 \\
  & \theta_{00} = 0.5 & \hspace*{5em} }}
\vspace{-3em}
\caption{An example prediction suffix tree}\label{fig:pst}
\end{figure}
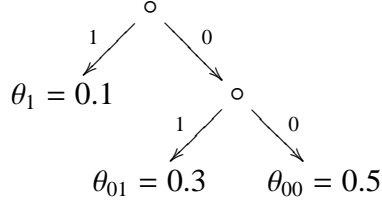

In practice, to use prediction suffix trees for binary sequence prediction, we need to learn both the model and parameter of a prediction
suffix tree from data.
We will deal with the model-learning part later.
Assuming the model of a PST is known/given, the parameter of the PST can be learnt using the KT estimator as follows.
We start with $\theta_l := \Pr_{kt}(1 \cbar \epsilon) = 1/2$ at each leaf node $l$ of $M$.
If $d$ is the depth of $M$, then the first $d$ bits $y_{1:d}$ of the input sequence are set aside for use as an initial context and
the variable $h$ denoting the bit sequence seen so far is set to $y_{1:d}$.
We then repeat the following steps as long as needed:
\begin{enumerate}\itemsep1mm\parskip0mm
 \item predict the next bit using the distribution $\theta_{M(h)}$;
 \item observe the next bit $y$, update $\theta_{M(h)}$ using Formula (\ref{kt update}) by incrementing either $a$ or $b$ according to the value
       of $y$, and then set $h := hy$.
\end{enumerate}

\paradot{Action-conditional PST}
The above describes how a PST is used for binary sequence prediction.
In the agent setting, we reduce the problem of predicting history sequences with general non-binary alphabets to that of predicting the
bit representations of those sequences.
Furthermore, we only ever condition on actions.
This is achieved by appending bit representations of actions to the input sequence without a corresponding update of the KT estimators.
These ideas are now formalised.

For convenience, we will assume without loss of generality that $|{\cal A}| = 2^{l_{\cal A}}$ and $|{\cal X}| = 2^{l_{\cal X}}$ for some $l_{\cal A},l_{\cal X} > 0$.
Given $a \in {\cal A}$, we denote by $\bstr{a} = a[1,l_{\cal A}] = a[1]a[2]\ldots a[l_{\cal A}] \in \{0,1\}^{l_{\cal A}}$ the bit representation of $a$.
Observation and reward symbols are treated similarly.
Further, the bit representation of a symbol sequence $x_{1:t}$ is denoted by $\bstr{x_{1:t}} = \bstr{x_1}\bstr{x_2}\ldots\bstr{x_t}$.

To do action-conditional sequence prediction using a PST with a given model $M$, we again start with
$\theta_l := \Pr_{kt}(1 \cbar \epsilon) = 1/2$ at each leaf node $l$ of $M$.
We also set aside a sufficiently long initial portion of the binary history sequence corresponding to the first few cycles to initialise the variable $h$ as usual.
The following steps are then repeated as long as needed:
\begin{enumerate}\itemsep1mm\parskip0mm
 \item set $h := h\bstr{a}$, where $a$ is the current selected action;
 \item for $i := 1$ to $l_{\cal X}$ do
  \begin{enumerate}\itemsep1mm\parskip0mm
    \item predict the next bit using the distribution $\theta_{M(h)}$;
    \item observe the next bit $x[i]$, update $\theta_{M(h)}$ using Formula (\ref{kt update}) according to the value of $x[i]$, and then set $h := hx[i]$.
  \end{enumerate}
\end{enumerate}

Let $M$ be the model of a prediction suffix tree, $a_{1:t} \in {\cal A}^t$ an action sequence,
$x_{1:t} \in {\cal X}^t$ an observation-reward sequence, and $h := \bstr{ax_{1:t}}$.
For each node $n$ in $M$, define $h_{M,n}$ by
\begin{equation}\label{eq:bits at a node} h_{M,n} := h_{i_1} h_{i_2} \cdots h_{i_k}\end{equation}
where $1 \leq i_1 < i_2 < \cdots < i_k \leq t$ and, for each $i$, $i \in \{i_1,i_2,\ldots i_k\} \text{\;iff\;}
h_{i}$ is an observation-reward bit and $n$ is a prefix of $M(h_{1:i-1})$.
In other words, $h_{M,n}$ consists of all the observation-reward bits with context $n$.
Thus we have the following expression for the probability of $x_{1:t}$ given $M$ and $a_{1:t}$:
\begin{align}
 \Pr( x_{1:t} \cdbar M, a_{1:t} )
  &= \prod_{i=1}^t \Pr( x_i \cbar M, ax_{<i}a_i) \notag \\
  &=  \prod_{i=1}^t \prod_{j=1}^{l_{\cal X}} \Pr( x_i[j] \cbar M, \bstr{ax_{<i}a_i} x_i[1,j-1]) \notag \\
  &= \prod_{n \in L(M)} \text{Pr$_{kt}$} ( h_{M,n}). \label{tree prop1}
\end{align}

The last step follows by grouping the individual probability terms according to the node $n \in L(M)$ in which each bit falls and then observing Equation (\ref{kt block}).
The above deals with action-conditional prediction using a single PST.
We now show how we can perform efficient action-conditional prediction using a Bayesian mixture of PSTs.
First we specify a prior over PST models.

\paradot{A Prior on Models of PSTs}
Our prior $\Pr(M) := 2^{-\Gamma_D(M)}$ is derived from a natural prefix coding of the tree structure of a PST.
The coding scheme works as follows: given a model of a PST of maximum depth $D$, a pre-order traversal of the tree is performed.
Each time an internal node is encountered, we write down 1.
Each time a leaf node is encountered, we write a 0 if the depth of the leaf node is less than $D$; otherwise we write nothing.
For example, if $D = 3$, the code for the model shown in Figure \ref{fig:pst} is 10100; if $D = 2$, the code for the same model is 101.
The cost $\Gamma_{D}(M)$ of a model $M$ is the length of its code, which is given by the number of nodes in $M$ minus
the number of leaf nodes in $M$ of depth $D$.
One can show that
\[ \sum_{M \in C_D} 2^{-\Gamma_{D}(M)} = 1, \]
where $C_D$ is the set of all models of prediction suffix trees with depth at most $D$; i.e.\ the prefix code is complete.
We remark that the above is another way of describing the coding scheme in \citeA{ctw95}.
Note that this choice of prior imposes an Ockham-like penalty on large PST structures.

\paradot{Context Trees}
The following data structure is a key ingredient of the Action-Conditional CTW algorithm.
\begin{defn}\label{defn:context tree}
A context tree of depth $D$ is a perfect binary tree of depth $D$ such that
attached to each node (both internal and leaf) is a probability on $\{0,1\}^*$.
\end{defn}

The node probabilities in a context tree are estimated from data by using a KT estimator at each node.
The process to update a context tree with a history sequence is similar to a PST, except that:
\begin{enumerate}\itemsep1mm\parskip0mm
 \item the probabilities at each node in the path from the root to a leaf traversed by an observed bit are updated; and
 \item we maintain block probabilities using Equations (\ref{kt block}) to (\ref{kt block inc}) instead of conditional probabilities.
\end{enumerate}
This process can be best understood with an example.
Figure~\ref{fig:ct} (left) shows a context tree of depth two.
For expositional reasons, we show binary sequences at the nodes; the node probabilities are computed from these.
Initially, the binary sequence at each node is empty.
Suppose $1001$ is the history sequence.
Setting aside the first two bits 10 as an initial context, the tree in the middle of Figure~\ref{fig:ct} shows what we have after
processing the third bit 0.
The tree on the right is the tree we have after processing the fourth bit 1.
In practice, we of course only have to store the counts of zeros and ones instead of complete subsequences at each node because, as we saw earlier in
(\ref{kt block}), $\Pr_{kt}(s) = \Pr_{kt}(a_s,b_s)$.
Since the node probabilities are completely determined by the input sequence, we shall henceforth speak unambiguously about {\em the} context
tree after seeing a sequence.

\begin{figure}
\hspace{0em}
\centerline{
\xymatrix @ur {
\epsilon   & \epsilon \ar[l]_1 \ar[d]^0 & \epsilon \ar[l]_1 \ar[d]^0 \\
& \epsilon \;\; \epsilon & \epsilon \ar[l]_1 \ar[d]^0 \\
&  & \epsilon
}
\hspace{0.25em}
\xymatrix @ur {
\epsilon   & \epsilon \ar[l]_1 \ar[d]^0 & 0 \ar[l]_1 \ar[d]^0 \\
& \epsilon \;\; 0 & 0 \ar[l]_1 \ar[d]^0 \\
&  & \epsilon
}
\hspace{0.25em}
\xymatrix @ur {
\epsilon   & \epsilon \ar[l]_1 \ar[d]^0 & 01 \ar[l]_1 \ar[d]^0 \\
& \epsilon \;\; 0 & 01 \ar[l]_1 \ar[d]^0 \\
&  & 1
}}
\vspace{-5.5em}
\caption{A depth-2 context tree (left); trees after processing two bits (middle and right)}\label{fig:ct}
\end{figure}
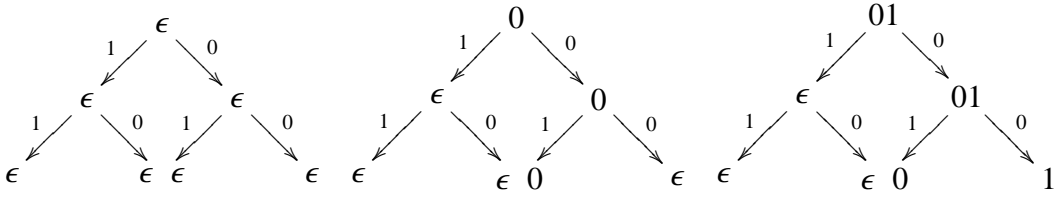

The context tree of depth $D$ after seeing a sequence $h$ has the following important properties:
\begin{enumerate}\itemsep1mm\parskip0mm
 \item the model of every PST of depth at most $D$ can be obtained from the context tree by pruning off appropriate subtrees and treating them as leaf nodes;
 \item the block probability of $h$ as computed by each PST of depth at most $D$ can be obtained from the node probabilities of the context tree via
       Equation~(\ref{tree prop1}).
\end{enumerate}
These properties, together with an application of the distributive law, form the basis of the highly efficient Action Conditional CTW algorithm.
We now formalise these insights.

\paradot{Weighted Probabilities}
The weighted probability $P^n_w$ of each node $n$ in the context tree $T$ after seeing $h := \bstr{ax_{1:t}}$ is defined inductively as follows:
\begin{align}\label{eq:rec_weighted_prob_update}
P^n_w := \begin{cases}
                                 \Pr_{kt} ( h_{T,n}) & \text{if $n$ is a leaf node;} \\
                                 \frac{1}{2}\Pr_{kt}( h_{T,n}) +
                                  \frac{1}{2}P^{n0}_w \times P^{n1}_w & \text{otherwise,}
                                \end{cases}
\end{align}
where $h_{T,n}$ is as defined in (\ref{eq:bits at a node}).

\begin{lem}[\citeA{ctw95}]
Let $T$ be the depth-$D$ context tree after seeing $h := \bstr{ax_{1:t}}$.
For each node $n$ in $T$ at depth $d$, we have
\begin{equation}
P^n_w  = \sum_{M \in C_{D-d}} 2^{-\Gamma_{D-d}(M)} \prod_{n' \in L(M)} \text{{\em Pr}}_{kt}( h_{T,nn'}).
\end{equation}
\label{model averaging}
\end{lem}
\newcommand{\half}{\frac{1}{2}}
\newcommand{\prM}{2^{-\Gamma_{\overline{d+1}}(M)}}
\newcommand{\rootterm}{\text{Pr$_{kt}$}( h_{T,n})}
\newcommand{\cterm}[2]{\prod_{n' \in L(#1)} \text{Pr}_{kt}( h_{T,#2n'})}
\begin{proof}
{\allowdisplaybreaks
The proof proceeds by induction on $d$.
The statement is clearly true for the leaf nodes at depth $D$.
Assume now the statement is true for all nodes at depth $d+1$, where $0 \leq d < D$.
Consider a node $n$ at depth $d$.
Letting $\overline{d} = D - d$, we have
\begin{align*}
P^n_w
 &= \half\rootterm + \half P^{n0}_w P^{n1}_w \notag\\
 &= \half\rootterm + \half \left[ \sum_{M \in C_{\overline{d+1}}} \prM \cterm{M}{n0} \right]
                                              \left[ \sum_{M \in C_{\overline{d+1}}} \prM \cterm{M}{n1} \right] \notag\\
 &= \half\rootterm \;+ \sum_{M_1 \in C_{\overline{d+1}}}\sum_{ M_2 \in C_{\overline{d+1}}}  2^{-(\Gamma_{\overline{d+1}}(M_1) + \Gamma_{\overline{d+1}}(M_2) +1)}
 \left[ \cterm{M_1}{n0} \right] \left[ \cterm{M_2}{n1} \right]\\
 &= \half\rootterm + \sum_{ \widehat{M_1M_2} \in C_{\overline{d}}} 2^{ -\Gamma_{\overline{d}}(\widehat{M_1M_2}) } \cterm{\widehat{M_1M_2}}{n}\\
 &= \sum_{ M \in C_{D-d}} 2^{-\Gamma_{D-d}(M)} \cterm{M}{n}, \notag
\end{align*}
where $\widehat{M_1M_2}$ denotes the tree in $C_{\overline{d}}$
whose left and right subtrees are $M_1$ and $M_2$ respectively.
}
\end{proof}

\paradot{Action Conditional CTW as a Mixture Environment Model}
A corollary of Lemma \ref{model averaging} is that at the root node $\epsilon$ of the context tree $T$ after seeing $h := \bstr{ax_{1:t}}$,
we have
{\allowdisplaybreaks
\begin{align}
P^\epsilon_w
  &= \sum_{ M \in C_D} 2^{-\Gamma_{D}(M)} \prod_{l \in L(M)} \text{Pr}_{kt}( h_{T,l}) \label{corr 1}\\
  &= \sum_{ M \in C_D} 2^{-\Gamma_{D}(M)} \prod_{l \in L(M)} \text{Pr}_{kt}( h_{M,l}) \label{corr 2} \\
  &= \sum_{ M \in C_D} 2^{-\Gamma_{D}(M)} \Pr( x_{1:t} \cdbar M, a_{1:t}), \label{corr 3}
\end{align}
where} the last step follows from Equation (\ref{tree prop1}).
Equation (\ref{corr 3}) shows that the quantity computed by the Action-Conditional CTW algorithm is exactly a mixture environment model.
Note that the conditional probability is always defined, as CTW assigns a non-zero probability to any sequence.
To sample from this conditional probability, we simply sample the individual bits of $x_t$ one by one.

In summary, to do prediction using Action-Conditional CTW, we set aside a sufficiently long initial portion of the binary history sequence
corresponding to the first few cycles to initialise the variable $h$ and then repeat the following steps as long as needed:
\begin{enumerate}\itemsep1mm\parskip0mm
 \item set $h := h\bstr{a}$, where $a$ is the current selected action;
 \item for $i := 1$ to $l_{\cal X}$ do
  \begin{enumerate}\itemsep1mm\parskip0mm
    \item predict the next bit using the weighted probability $P^\epsilon_w$;
    \item observe the next bit $x[i]$, update the context tree using $h$ and $x[i]$, calculate the new weighted probability $P^\epsilon_w$, and then set $h := hx[i]$.
  \end{enumerate}
\end{enumerate}

\paradot{Incorporating Type Information}
One drawback of the Action-Conditional CTW algorithm is the potential loss of type information when mapping a history string to its binary encoding.
This type information may be needed for predicting well in some domains.
Although it is always possible to choose a binary encoding scheme so that the type information can be inferred by a depth limited context tree, it would be desirable to remove this restriction so that our agent can work with arbitrary encodings of the percept space.

One option would be to define an action-conditional version of multi-alphabet CTW \cite{tjalkens93}, with the alphabet consisting of the entire percept space.
The downside of this approach is that we then lose the ability to exploit the structure within each percept.
This can be critical when dealing with large observation spaces, as noted by \citeA{mccallum96}.
The key difference between his U-Tree and USM algorithms is that the former could discriminate between individual components within an observation, whereas the latter worked only at the symbol level.
As we shall see in Section \ref{sec:experiments}, this property can be helpful when dealing with larger problems.

Fortunately, it is possible to get the best of both worlds.
We now describe a technique that incorporates type information whilst still working at the bit level.
The trick is to chain together $k:=l_{\cal X}$ action conditional PSTs, one for each bit of the percept space, with appropriately overlapping binary contexts.
More precisely, given a history $h$, the context for the $i$'th PST is the most recent $D+i-1$ bits of the bit-level history string $\bstr{h}x[1,i-1]$.
To ensure that each percept bit is dependent on the same portion of $h$, $D + i - 1$ (instead of only $D$) bits are used.
Thus if we denote the PST model for the $i$th bit in a percept $x$ by $M_i$, and the joint model by $M$, we now have:
\begin{align}
\Pr(x_{1:t} \cdbar M, a_{1:t})
  &= \prod_{i=1}^t \Pr( x_i \cbar M, ax_{<i}a_i) \notag \\
  &= \prod_{i=1}^t \prod_{j=1}^{k} \Pr( x_i[j] \cbar M_j, \bstr{ax_{<i}a_i} x_i[1,j-1]) \label{eq:fac middle} \\
  &= \prod_{j=1}^{k} \Pr( x_{1:t}[j] \cbar M_j, x_{1:t}[-j], a_{1:t}) \notag
\end{align}
where $x_{1:t}[i]$ denotes $x_1[i]x_2[i] \ldots x_t[i]$, $x_{1:t}[-i]$ denotes $x_1[-i] x_2[-i] \ldots x_t[-i]$, with $x_t[-j]$ denoting $x_t[1] \ldots x_t[j-1] x_t[j+1] \ldots x_t[k]$.
The last step follows by swapping the two products in (\ref{eq:fac middle}) and using the above notation to refer to the product of probabilities of the $j$th bit in each percept $x_i$, for $1 \leq i \leq t$.

We next place a prior on the space of factored PST models $M \in C_{D} \times \dots \times C_{D+k-1}$ by assuming that each factor is independent, giving
\begin{equation*}
\Pr(M) = \Pr(M_1, \dots, M_{k}) = \prod_{i=1}^{k}{2^{-\Gamma_{D_i}(M_i)}} = 2^{-\sum\limits_{i=1}^k{\Gamma_{D_i}(M_i)}},
\end{equation*}
where $D_i := D+i-1$.
This induces the following mixture environment model
\begin{equation}\label{eq:factored_ctw}
\xi( x_{1:t} \cbar a_{1:t}) := \sum_{ M \in C_{D_1} \times \dots \times C_{D_k}} 2^{-\sum\limits_{i=1}^{k}{\Gamma_{D_i}(M_i)}} \Pr( x_{1:t} \cbar M, a_{1:t}).
\end{equation}
This can now be rearranged into a product of efficiently computable mixtures, since
\begin{eqnarray}
\xi( x_{1:t} \cbar a_{1:t})  &=&  \sum_{ M_1 \in C_{D_1}} \dots \sum_{ M_k \in C_{D_k}}  2^{- \sum\limits_{i=1}^k{\Gamma_{D_i}(M_i)}} \prod_{j=1}^k \Pr( x_{1:t}[j] \cdbar M_j, x_{1:t}[-j], a_{1:t}) \notag\\
&=& \prod_{j=1}^k \left( \sum_{ M_j \in C_{D_j}} 2^{-\Gamma_{D_j}(M_j)} \Pr( x_{1:t}[j] \cdbar M_j, x_{1:t}[-j], a_{1:t}) \right). \label{eq:fac_ctw}
\end{eqnarray}
Note that for each factor within Equation (\ref{eq:fac_ctw}), a result analogous to Lemma \ref{model averaging} can be established by appropriately modifying Lemma \ref{model averaging}'s proof to take into account that now only one bit per percept is being predicted.
This leads to the following scheme for incrementally maintaining Equation (\ref{eq:factored_ctw}):

\begin{enumerate}\itemsep1mm\parskip0mm
\item Initialise $h \leftarrow \epsilon$, $t \leftarrow 1$. Create $k$ context trees.
\item Determine action $a_t$. Set $h \leftarrow ha_t$.
\item Receive $x_t$. For each bit $x_t[i]$ of $x_t$, update the $i$th context tree with $x_t[i]$ using history \text{$hx[1,i-1]$} and recompute $P^\epsilon_w$ using Equation (\ref{eq:rec_weighted_prob_update}).
\item  Set $h \leftarrow hx_t$, $t \leftarrow t+1$. Goto $2$.
\end{enumerate}
We will refer to this technique as Factored Action-Conditional CTW, or the {\sc \text{FAC-CTW}} algorithm for short.

\paradot{Convergence to the True Environment}
We now show that \predictor\ performs well in the class of stationary $n$-Markov environments.
Importantly, this includes the class of Markov environments used in state-based reinforcement learning, where the most recent action/observation pair $(a_t, x_{t-1})$ is a sufficient statistic for the prediction of $x_t$.

\begin{defn}
Given $n \in \mathbb{N}$, an environment $\mu$ is said to be $n$-Markov if
for all $t>n$, for all $a_{1:t} \in \cA^t$, for all $x_{1:t} \in \cX^t$ and for all $h \in (\cA \times \cX)^{t-n-1} \times \cA$
\begin{equation}
\mu(x_t \cbar ax_{<t}a_t) = \mu(x_t \cbar h x_{t-n} ax_{t-n+1:t-1} a_t).
\end{equation}
Furthermore, an $n$-Markov environment is said to be stationary if for all $ax_{1:n} a_{n+1} \in (\cA \times \cX)^n \times \cA$,  for all $h,h' \in (\cA \times \cX)^*$,
\begin{equation}\label{eq:stationary}
 \mu( \cdot \cbar h ax_{1:n} a_{n+1}) = \mu( \cdot \cbar h'ax_{1:n} a_{n+1}).
\end{equation}
\end{defn}

It is easy to see that any stationary $n$-Markov environment can be represented as a product of sufficiently large, fixed parameter PSTs.
Theorem \ref{conv to mu} states that the predictions made by a mixture environment model only converge to those of the true environment when the model class contains a model sufficiently close to the true environment.
However, no \emph{stationary} \text{$n$-Markov} environment model is contained within the model class of \predictor, since each model updates the parameters for its KT-estimators as more data is seen.
Fortunately, this is not a problem, since this updating produces models that are sufficiently close to any stationary $n$-Markov environment for Theorem \ref{conv to mu} to be meaningful.

\begin{lem}\label{lem:redundancy_bound}
If $\mathcal{M}$ is the model class used by \predictor\ with a context depth
$D$,
$\mu$ is an environment expressible as a product of $k:=l_{\cX}$ fixed parameter PSTs $(M_1, \Theta_1), \dots, (M_k, \Theta_k)$ of maximum depth $D$ and
$\rho(\cdot \cbar a_{1:n}) \equiv \Pr( \cdot \cbar (M_1,\ldots,M_k), a_{1:n}) \in {\cal M}$
then for all $n \in \mathbb{N}$, for all $a_{1:n} \in \cA^n$,
\begin{equation*}
D_{1:n}(\mu ~||~ \rho) \leq  \sum\limits_{j=1}^k |L(M_j)| ~ \gamma \left(\frac{n}{|L(M_j)|} \right)
\end{equation*}
where
\begin{equation*}
\gamma(z):=  \left\{
     \begin{array}{lll}
       z & \text{ for } & 0 \leq z < 1\\
       \frac{1}{2} \log z + 1 & \text{ for } & z \geq 1.
     \end{array}
   \right.
\end{equation*}
\begin{proof}
For all $n \in \mathbb{N}$, for all $a_{1:n} \in \cA^n$,
\begin{align}
D_{1:n}(\mu ~||~ \rho)
= & \sum\limits_{x_{1:n}} \mu(x_{1:n} \cbar a_{1:n}) \ln \frac{\mu(x_{1:n} \cbar a_{1:n})}{\rho(x_{1:n} \cbar a_{1:n})} \notag \\
= & \sum\limits_{x_{1:n}} \mu(x_{1:n} \cbar a_{1:n}) \ln \frac{\prod_{j=1}^{k} \Pr( x_{1:n}[j] \cbar M_j, \Theta_j, x_{1:n}[-j], a_{1:n})}{\prod_{j=1}^{k} \Pr( x_{1:n}[j] \cbar M_j, x_{1:n}[-j], a_{1:n})} \notag \\
= & \sum\limits_{x_{1:n}}  \mu(x_{1:n} \cbar a_{1:n}) \sum\limits_{j=1}^{k} \ln \frac{\Pr( x_{1:n}[j] \cbar M_j, \Theta_j, x_{1:n}[-j], a_{1:n})}{\Pr( x_{1:n}[j] \cbar M_j, x_{1:n}[-j], a_{1:n})} \notag \\
\leq& \sum\limits_{x_{1:n}}  \mu(x_{1:n} \cbar a_{1:n}) \sum\limits_{j=1}^k |L(M_j)| \gamma \left(\frac{n}{|L(M_j)|} \right) \label{step:redundancy_bound} \\
= & ~ \sum\limits_{j=1}^k |L(M_j)| ~ \gamma \left(\frac{n}{|L(M_j)|} \right) \notag
\end{align}
where $\Pr( x_{1:n}[j] \cbar M_j, \Theta_j, x_{1:n}[-j], a_{1:n})$ denotes the probability of a fixed parameter PST $(M_j, \Theta_j)$ generating the sequence $x_{1:n}[j]$ and the bound introduced in (\ref{step:redundancy_bound}) is from \cite{ctw95}.
\end{proof}
\end{lem}

If the unknown environment $\mu$ is stationary and $n$-Markov, Lemma \ref{lem:redundancy_bound} and Theorem \ref{conv to mu} can be applied to the \predictor\ mixture environment model $\xi$.
Together they imply that the cumulative $\mu$-expected squared difference between $\mu$ and $\xi$ is bounded by $O(\log n)$.
Also, the \emph{per cycle} $\mu$-expected squared difference between $\mu$ and $\xi$ goes to zero at the rapid rate of $O(\log n/n)$.
This allows us to conclude that \predictor\ (with a sufficiently large context depth) will perform well on the class of stationary $n$-Markov environments.

\paradot{Summary}
We have described two different ways in which CTW can be extended to define a large and efficiently computable mixture environment model.
The first is a complete derivation of the Action-Conditional CTW algorithm first presented in \cite{veness10}.
The second is the introduction of the {\sc \text{FAC-CTW}} algorithm, which improves upon Action-Conditional CTW by automatically exploiting the type information available within the agent setting.

As the rest of the paper will make extensive use of the FAC-CTW algorithm, for clarity we define
\begin{equation}\label{eq:facctw_mixture}
\Upsilon( x_{1:t} \cbar a_{1:t}) := \sum_{ M \in C_{D_1} \times \dots \times C_{D_k}} \hspace{-1.2em} 2^{-\sum\limits_{i=1}^k{\Gamma_{D_i}(M_i)}} \Pr( x_{1:t} \cbar M, a_{1:t}).
\end{equation}
Also recall that using $\Upsilon$ as a mixture environment model, the conditional probability of $x_{t}$ given $ax_{<t}a_t$ is
\[ \Upsilon( x_t \cbar ax_{<t}a_t) = \frac{ \Upsilon( x_{1:t} \cdbar a_{1:t}) }{ \Upsilon( x_{<t} \cdbar a_{<t}) }, \]
which follows directly from Equation (\ref{cond prob of or}).
To generate a percept from this conditional probability distribution, we simply sample $l_{\cX}$ bits, one by one, from $\Upsilon$.

\paradot{Relationship to AIXI}
Before moving on, we examine the relationship between AIXI and our model class approximation.
{\allowdisplaybreaks
Using $\Upsilon$ in place of $\rho$ in Equation (\ref{value defn}), the optimal action for an agent at time $t$, having experienced
$ax_{1:t-1}$, is given by
\begin{align}
a_t^*
=& \arg \max\limits_{a_t}\sum\limits_{x_t} \frac{ \Upsilon( x_{1:t} \cdbar a_{1:t}) }{ \Upsilon( x_{<t} \cdbar a_{<t})} \cdots
\max\limits_{a_{t+m}}\sum\limits_{x_{t+m}} \frac{ \Upsilon( x_{1:t+m} \cdbar a_{1:t+m} ) } { \Upsilon( x_{<t+m} \cdbar a_{<t+m}) }
\left[\sum\limits_{i=t}^{t+m}r_i\right] \notag\\
= &  \arg \max\limits_{a_t}\sum\limits_{x_t} \cdots
\max\limits_{a_{t+m}}\sum\limits_{x_{t+m}} \left[ \sum\limits_{i=t}^{t+m}r_i \right] \prod_{i=t}^{t+m} \frac{ \Upsilon ( x_{1:i} \cdbar a_{1:i}) } {
  \Upsilon ( x_{<i} \cdbar a_{<i}) } \notag\\
= &  \arg \max\limits_{a_t}\sum\limits_{x_t}\cdots \max\limits_{a_{t+m}}\sum\limits_{x_{t+m}} \left[ \sum\limits_{i=t}^{t+m}r_i \right] \frac{ \Upsilon( x_{1:t+m} \cdbar a_{1:t+m}) }{\Upsilon(x_{<t} \cdbar a_{<t})}\notag\\
= &  \arg \max\limits_{a_t}\sum\limits_{x_t}\cdots \max\limits_{a_{t+m}}\sum\limits_{x_{t+m}} \left[ \sum\limits_{i=t}^{t+m}r_i \right] \Upsilon( x_{1:t+m} \cdbar a_{1:t+m})  \notag\\
= & \arg \max\limits_{a_t}\sum\limits_{x_t}\cdots \max\limits_{a_{t+m}}\sum\limits_{x_{t+m}} \left[ \sum\limits_{i=t}^{t+m}r_i \right] \sum_{ M \in C_{D_1} \times \dots \times C_{D_k}} \hspace{-1.2em} 2^{-\sum\limits_{i=1}^k{\Gamma_{D_i}(M_i)}} \Pr( x_{1:t+m} \cbar M, a_{1:t+m}). \label{aipsi}
\end{align}
}
Contrast (\ref{aipsi}) now with Equation (\ref{aixi_eq2}) which we reproduce here:
\begin{equation}
a_t^* = \arg\max\limits_{a_t}\sum\limits_{x_t} \dots \max\limits_{a_{t+m}} \sum\limits_{x_{t+m}}
\left[\sum\limits_{i=t}^{t+m}r_i\right]
\sum\limits_{\rho \in {\cal M}}2^{-K(\rho)}\rho(x_{1:t+m} \cdbar a_{1:t+m}),
\end{equation}
where ${\cal M}$ is the class of all enumerable chronological semimeasures, and $K(\rho)$ denotes the Kolmogorov complexity of $\rho$.
The two expressions share a prior that enforces a bias towards simpler models.
The main difference is in the subexpression describing the mixture over the model class.
AIXI uses a mixture over all enumerable chronological semimeasures.
This is scaled down to a (factored) mixture of prediction suffix trees in our setting.
Although the model class used in AIXI is completely general, it is also incomputable.
Our approximation has restricted the model class to gain the desirable computational properties of {\sc FAC-CTW}.

\section{Putting it All Together}\label{sec:together}

Our approximate AIXI agent, \agent, is realised by instantiating the \searchalg\ algorithm with $\rho=\Upsilon$.
Some additional properties of this combination are now discussed.

\paradot{Convergence of Value}
We now show that using $\Upsilon$ in place of the true environment $\mu$ in the expectimax operation leads to good behaviour when $\mu$ is both stationary and $n$-Markov.
This result combines Lemma \ref{lem:redundancy_bound} with an adaptation of \cite[Thm.5.36]{Hutter:04uaibook}.
For this analysis, we assume that the instantaneous rewards are non-negative (with no loss of generality), \predictor\ is used with a sufficiently large context depth, the maximum life of the agent $b\in\mathbb{N}$ is fixed and that a bounded planning horizon $m_t := \min(H, b-t+1)$ is used at each time $t$, with $H \in \mathbb{N}$ specifying the maximum planning horizon.
\begin{thm}\label{thm:value_conv}
Using the \predictor\ algorithm, for every policy $\pi$, if the true environment $\mu$ is expressible as a product of $k$ PSTs $(M_1, \Theta_1), \dots, (M_k, \Theta_k)$, for all $b\in \mathbb{N}$, we have
\begin{equation*}
\sum\limits_{t=1}^{b} \mathbb{E}_{x_{<t} \sim \mu} \left[ \left( v^{m_t}_\Upsilon(\pi, ax_{<t}) - v^{m_t}_{\mu}(\pi, ax_{<t}) \right)^2 \right]
\leq 2 H^3 r_{max}^2 \left[ \sum\limits_{i=1}^k{\Gamma_{D_i}(M_i)} + \sum\limits_{j=1}^k |L(M_j)| ~ \gamma \left(\frac{b}{|L(M_j)|} \right)   \right]
\end{equation*}
where $r_{max}$ is the maximum instantaneous reward, $\gamma$ is as defined in Lemma~\ref{lem:redundancy_bound} and $v^{m_t}_{\mu}(\pi, ax_{<t})$ is the value of policy $\pi$ as defined in Definition \ref{defn:policy_value}.
\begin{proof}
First define $\rho(x_{i:j} \cdbar a_{1:j}, x_{<i}) := \rho(x_{1:j} \cdbar a_{1:j}) / \rho(x_{<i} \cdbar a_{<i})$ for $i<j\;$, for any environment model $\rho$ and let $a_{t:m_t}$ be the actions chosen by $\pi$ at times $t$ to $m_t$.
Now
\begin{eqnarray*}
\left| v^{m_t}_\Upsilon(\pi, ax_{<t}) - v^{m_t}_{\mu}(\pi, ax_{<t}) \right| &=& \left| \sum_{x_{t:m_t}} (r_t + \dots + r_{m_t}) \left[ \Upsilon(x_{t:m_t} \cdbar a_{1:m_t}, x_{<t}) - \mu(x_{t:m_t} \cdbar a_{1:m_t}, x_{<t}) \right] \right|\\
&\leq& \sum_{x_{t:m_t}} (r_t + \dots + r_{m_t}) \left| \Upsilon(x_{t:m_t} \cdbar a_{1:m_t}, x_{<t}) - \mu(x_{t:m_t} \cdbar a_{1:m_t}, x_{<t}) \right| \\
&\leq& m_t r_{max} \sum_{x_{t:m_t}} \left| \Upsilon(x_{t:m_t} \cdbar a_{1:m_t}, x_{<t}) - \mu(x_{t:m_t} \cdbar a_{1:m_t}, x_{<t}) \right| \\
&=:& m_t r_{max} A_{t:m_t}(\mu \;||\; \Upsilon).
\end{eqnarray*}
Applying this bound, a property of absolute distance \cite[Lemma 3.11]{Hutter:04uaibook} and the chain rule for KL-divergence \cite[p. 24]{coverthomas} gives
\begin{gather*}
\sum_{t=1}^b \mathbb{E}_{x_{<t} \sim \mu} \left[ \left( v^{m_t}_\Upsilon(\pi, ax_{<t}) - v^{m_t}_{\mu}(\pi, ax_{<t}) \right)^2 \right] \leq
m_t^2 r_{max}^2 \sum_{t=1}^b \mathbb{E}_{x_{<t} \sim \mu} \left[ A_{t:m_t}(\mu \;||\; \Upsilon)^2 \right]\\
\hspace{2em} \leq 2 H^2 r_{max}^2 \sum_{t=1}^b \mathbb{E}_{x_{<t} \sim \mu} \left[ D_{t:m_t}(\mu \;||\; \Upsilon) \right]
= 2 H^2 r_{max}^2 \sum_{t=1}^b \sum_{i=t}^{m_t} \mathbb{E}_{x_{<i} \sim \mu} \left[ D_{i:i}(\mu \;||\; \Upsilon) \right]\\
\hspace{2em} \leq 2 H^3 r_{max}^2 \sum_{t=1}^{b} \mathbb{E}_{x_{<t} \sim \mu} \left[ D_{t:t}(\mu \;||\; \Upsilon) \right]
= 2 H^3 r_{max}^2 D_{1:b}(\mu \;||\; \Upsilon),
\end{gather*}
where $D_{i:j}(\mu \;||\; \Upsilon) := \sum_{x_{i:j}}  \mu(x_{i:j} \cdbar a_{1:j}, x_{<i}) \ln (\Upsilon(x_{i:j} \cdbar a_{1:j}, x_{<i})/ \mu(x_{i:j} \cdbar a_{1:j}, x_{<i}))$.
The final inequality uses the fact that any particular $D_{i:i}(\mu \;||\; \Upsilon)$ term appears at most $H$ times in the preceding double sum.
Now define $\rho_M(\cdot \cdbar a_{1:b}) := \Pr( \cdot \cbar (M_1,\ldots,M_k), a_{1:b})$ and we have
\begin{eqnarray*}
D_{1:b}(\mu \;||\; \Upsilon)
&=& \sum_{x_{1:b}} \mu( x_{1:b} \cdbar a_{1:b}) \ln \left[ \frac{\mu(x_{1:b}\cdbar a_{1:b})}{\rho_M( x_{1:b} \cdbar a_{1:b})}  \frac{\rho_M( x_{1:b} \cdbar
    a_{1:b})} {\Upsilon( x_{1:b} \cdbar a_{1:b})} \right] \\
 &=& \sum_{x_{1:b}} \mu( x_{1:b} \cdbar a_{1:b}) \ln \frac{\mu(x_{1:b}\cdbar a_{1:b})}{\rho_M( x_{1:b} \cdbar a_{1:b})} +
    \sum_{x_{1:b}} \mu( x_{1:b} \cdbar a_{1:b}) \ln \frac{\rho_M( x_{1:b} \cdbar
     a_{1:b})} {\Upsilon( x_{1:b} \cdbar a_{1:b})}\\
 &\leq& D_{1:b}( \mu \,\|\, \rho_M) + \sum_{x_{1:b}} \mu( x_{1:b} \cdbar a_{1:b}) \ln \frac{ \rho_M( x_{1:b} \cdbar a_{1:b}) }{ w_0^{\rho_M} \rho_M( x_{1:b} \cdbar a_{1:b}) } \\
 &=& D_{1:b}( \mu \,\|\, \rho_M) + \sum\limits_{i=1}^k{\Gamma_{D_i}(M_i)}
\end{eqnarray*}
where $w_0^{\rho_M} := 2^{-\sum\limits_{i=1}^k{\Gamma_{D_i}(M_i)}}$ and the final inequality follows by dropping all but $\rho_M$'s contribution to Equation (\ref{eq:facctw_mixture}).
Using Lemma \ref{lem:redundancy_bound} to bound $D_{1:b}( \mu \,\|\, \rho_M)$ now gives the desired result.
\end{proof}
\end{thm}
For any fixed $H$, Theorem \ref{thm:value_conv} shows that the cumulative expected squared difference of the true
and $\Upsilon$ values is bounded by a term that grows at the rate of $O(\log b)$.
The average expected squared difference of the two values then goes down to zero at the rate of $O(\frac{\log b}{b})$.
This implies that for sufficiently large $b$, the value estimates using $\Upsilon$ in place of $\mu$ converge for any fixed policy $\pi$.
Importantly, this includes the fixed horizon expectimax policy with respect to $\Upsilon$.

\paradot{Convergence to Optimal Policy}\label{sec:theory}
This section presents a result for $n$-Markov environments that are both ergodic and stationary.
Intuitively, this class of environments never allow the agent to make a mistake from which it can no longer recover.
Thus in these environments an agent that learns from its mistakes can hope to achieve a long-term average reward that will approach optimality.

\begin{defn}\label{def:ergodic_nmarkov_env}
An $n$-Markov environment $\mu$ is said to be ergodic if there exists a policy $\pi$ such that every sub-history $s \in \left(\cA \times \cX\right)^n$ possible in $\mu$ occurs infinitely often (with probability $1$) in the history generated by an agent/environment pair $\left(\pi, \mu\right)$.
\end{defn}

\begin{defn}
A sequence of policies $\left\{ \pi_1, \pi_2, \dots \right\}$ is said to be self optimising with respect to model class ${\cal M}$ if
\begin{equation}
\frac{1}{m} v^m_{\rho}(\pi_m, \epsilon) - \frac{1}{m} V^m_{\rho}(\epsilon) \to 0 \text{ \hspace{1em} as \hspace{1em}}  m \to \infty \text{\hspace{1em} for all \hspace{1em}} \rho \in \cal{M}.
\end{equation}
\end{defn}

A self optimising policy has the same long-term average expected future reward as the optimal policy for any environment in $\cal{M}$.
In general, such policies cannot exist for all model classes.
We restrict our attention to the set of stationary, ergodic $n$-Markov environments since these are what can be modeled effectively by \predictor.
The ergodicity property ensures that no possible percepts are precluded due to earlier actions by the agent.
The stationarity property ensures that the environment is sufficiently well behaved for a PST to learn a fixed set of parameters.

We now prove a lemma in preparation for our main result.

\begin{lem}\label{lem:nmarkov_to_mdp}
Any stationary, ergodic $n$-Markov environment can be modeled by a finite, ergodic MDP.
\begin{proof}
Given an ergodic $n$-Markov environment $\mu$, with associated action space $\cA$ and percept space $\cX$, an equivalent, finite MDP $(S, A, T, R)$ can be constructed from $\mu$ by defining the state space as $S :=\left(\cA \times \cX\right)^n$, the action space as $A:=\cA$, the transition probability as $T_a(s, s'):=\mu(o'r' \cbar hsa)$ and the reward function as $R_a(s, s'):=r'$, where $s'$ is the suffix formed by deleting the leftmost action/percept pair from $sao'r'$ and $h$ is an arbitrary history from $(\cA \times \cX)^*$.
$T_a(s, s')$ is well defined for arbitrary $h$ since $\mu$ is stationary, therefore Eq. (\ref{eq:stationary}) applies.
Definition \ref{def:ergodic_nmarkov_env} implies that the derived MDP is ergodic.
\end{proof}
\end{lem}

\begin{thm}\label{thm:self_optimising}
Given a mixture environment model $\xi$ over a model class ${\cal M}$ consisting of a countable set of stationary, ergodic $n$-Markov environments, the sequence of policies $\left\{ \pi^\xi_1, \pi^\xi_2, \dots \right\}$ where
\begin{equation}
\pi_b^\xi(ax_{<t}) := \arg\max_{a_t \in \cA} V_{\xi}^{b-t+1}(ax_{<t}a_t)
\end{equation}
for $1 \leq t \leq b$, is self-optimising with respect to model class $\cal{M}$.
\begin{proof}
By applying Lemma \ref{lem:nmarkov_to_mdp} to each $\rho \in {\cal M}$, an equivalent model class $\cal{N}$ of finite, ergodic MDPs can be produced.
We know from \citeA[Thm.5.38]{Hutter:04uaibook} that a sequence of policies for $\cal{N}$ that is self-optimising exists.
This implies the existence of a corresponding sequence of policies for $\cal{M}$ that is self-optimising.
Using \cite[Thm.5.29]{Hutter:04uaibook}, this implies that the sequence of policies $\left\{ \pi_1^\xi, \pi_2^\xi, \dots \right\}$ is self optimising.
\end{proof}
\end{thm}

Theorem \ref{thm:self_optimising} says that by choosing a sufficiently large lifespan $b$, the average reward for an agent following policy $\pi^\xi_b$ can be made arbitrarily close to the optimal average reward with respect to the true environment.

Theorem \ref{thm:self_optimising} and the consistency of the \searchalg\ algorithm (\ref{uctc}) give support to the claim that the \agent\ agent is self-optimising with respect to the class of stationary, ergodic, $n$-Markov environments.
The argument isn't completely rigorous, since the usage of the KT-estimator implies that the model class of \predictor\ contains an uncountable number of models.
Our conclusion is not entirely unreasonable however.
The justification is that a countable mixture of PSTs behaving similarly to the \predictor\ mixture can be formed by replacing each PST leaf node KT-estimator with a finely grained, discrete Bayesian mixture predictor.
Under this interpretation, a floating point implementation of the KT-estimator would correspond to a computationally feasible approximation of the above.

The results used in the proof of Theorem \ref{thm:self_optimising} can be found in \cite{Hutter:02selfopt} and \cite{Legg04ergodicmdps}.
An interesting direction for future work would be to investigate whether a self-optimising result similar to \cite[Thm.5.29]{Hutter:04uaibook} holds for continuous mixtures.

\paradot{Computational Properties}\label{subsec:comp considerations}
The \predictor\ algorithm grows each context tree data structure dynamically.
With a context depth $D$, there are at most $O(tD\log(|\cO||\cR|))$ nodes in the set of context trees after $t$ cycles.
In practice, this is considerably less than $\log(|\cO||\cR|) 2^D$, which is the number of nodes in a fully grown set of context trees.
The time complexity of \predictor\ is also impressive; $O(Dm\log(|\cO||\cR|))$ to generate the $m$ percepts
needed to perform a single \searchalg\ simulation and $O(D \log(|\cO||\cR|))$ to process each new piece of experience.
Importantly, these quantities are not dependent on $t$, which means that the performance of our agent does not degrade with time.
Thus it is reasonable to run our agent in an online setting for millions of cycles.
Furthermore, as {\sc FAC-CTW} is an exact algorithm, we do not suffer from approximation issues that plague sample based approaches to Bayesian learning.

\paradot{Efficient Combination of FAC-CTW with \searchalg}
Earlier, we showed how \predictor\ can be used in an online setting.
An additional property however is needed for efficient use within \searchalg.
Before {\sc Sample} is invoked, \predictor\ will have computed a set of context trees for a history of length $t$.
After a complete trajectory is sampled, \predictor\ will now contain a set of context trees for a history of length $t+m$.
The original set of context trees now needs to be restored.
Saving and copying the original context trees is unsatisfactory, as is rebuilding them from scratch in $O(tD\log(|\cO||\cR|))$ time.
Luckily, the original set of context trees can be recovered efficiently by traversing the history at time $t+m$ in reverse, and performing an inverse update operation on each of the $D$ affected nodes in the relevant context tree, for each bit in the sample trajectory.
This takes $O(Dm\log(|\cO||\cR|))$ time.
Alternatively, a copy on write implementation can be used to modify the context trees during the simulation phase, with the modified copies of each context node discarded before {\sc Sample} is invoked again.

\paradot{Exploration/Exploitation in Practice}\label{subsec:exploration exploitation in practice}
Bayesian belief updating combines well with expectimax based planning.
Agents using this combination, such as AIXI and \agent, will automatically perform information gathering actions if the expected reduction in uncertainty would lead to higher expected future reward.
Since AIXI is a mathematical notion, it can simply take a large initial planning horizon $b$, e.g. its maximal lifespan, and then
at each cycle $t$ choose greedily with respect to Equation~(\ref{aixi_eq}) using a \emph{remaining horizon} of $b - t + 1$.
Unfortunately in the case of \agent, the situation is complicated by issues of limited computation.

In theory, the \agent\ agent could always perform the action recommended by \searchalg.
In practice however, performing an expectimax operation with a remaining horizon of $b-t+1$ is not feasible, even using Monte-Carlo approximation.
Instead we use as large a fixed search horizon as we can afford computationally, and occasionally force exploration according to some heuristic policy.
The intuition behind this choice is that in many domains, good behaviour can be achieved by using a small amount of planning if the dynamics of the domain are known.
Note that it is still possible for \searchalg\ to recommend an exploratory action, but only if the benefits of this information can be realised within its limited planning horizon.
Thus, a limited amount of exploration can help the agent avoid local optima with respect to its present set of beliefs about the underlying environment.
Other online reinforcement learning algorithms such as {\sc SARSA($\lambda$)} \cite{sutton-barto98}, U-Tree \cite{mccallum96} or Active-LZ \cite{farias07} employ similar such strategies.

\begin{figure*}[!t]
\centerline{\mbox{\includegraphics[scale=1.1]{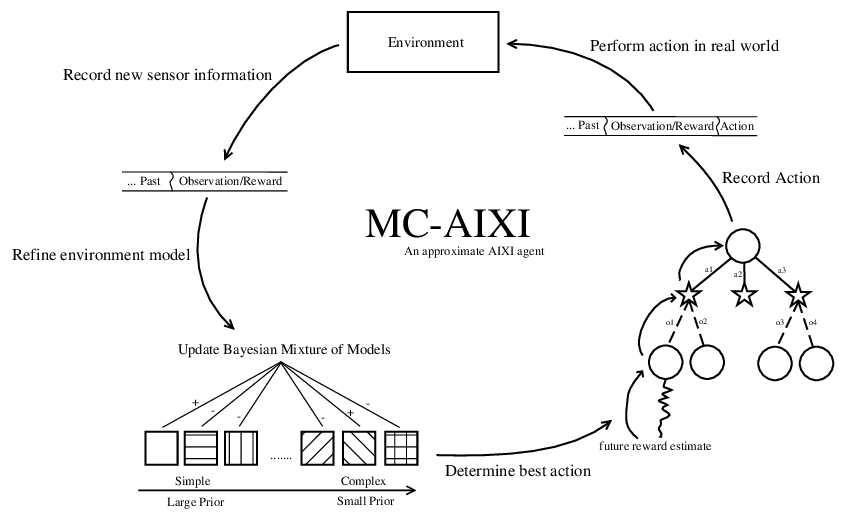}}}
\caption{The MC-AIXI agent loop}
\label{agent_overview}
\end{figure*}

\paradot{Top-level Algorithm}
At each time step, \agent\ first invokes the \searchalg\ routine with a fixed horizon to estimate the value of each candidate action.
An action is then chosen according to some policy that balances exploration with exploitation, such as $\epsilon$-Greedy or Softmax \cite{sutton-barto98}.
This action is communicated to the environment, which responds with an observation-reward pair.
The agent then incorporates this information into $\Upsilon$ using the \predictor\ algorithm and the cycle repeats.
Figure \ref{agent_overview} gives an overview of the agent/environment interaction loop.

\section{Experimental Results}\label{sec:experiments}

We now measure our agent's performance across a number of different domains.
In particular, we focused on learning and solving some well-known benchmark problems from the POMDP literature.
Given the full POMDP model, computation of the optimal policy for each of these POMDPs is not difficult.
However, our requirement of having to both learn a model of the environment, as well as find a good policy online, \emph{significantly} increases
the difficulty of these problems.
From the agent's perspective, our domains contain perceptual aliasing, noise, partial information, and inherent stochastic elements.

Our test domains are now described.
Their characteristics are summarized in Table~\ref{table:domain_characteristics}.
\begin{table}
\begin{center}
\begin{tabular}{|l|c|c|c|c|c|c|c|}
\hline
Domain & $|\cA|$ & $|\cO|$ & Aliasing & Noisy $\cO$ & Uninformative $\cO$\\
\hline
1d-maze        & 2 & 1  & yes & no & yes\\
Cheese Maze    & 4 & 16 & yes & no & no\\
Tiger          & 3 & 3  & yes & yes & no \\
Extended Tiger & 4 & 3  & yes & yes & no \\
4 $\times$ 4 Grid & 4 & 1 & yes & no & yes\\
TicTacToe & 9 & 19683 & no & no & no\\
Biased Rock-Paper-Scissor & 3 & 3 & no & yes & no\\
Kuhn Poker & 2 & 6 & yes & yes & no\\
Partially Observable Pacman & 4  & $2^{16}$ & yes  & no & no\\
\hline
\end{tabular}
\caption{Domain characteristics}
\label{table:domain_characteristics}
\end{center}
\end{table}
\paradot{1d-maze}
The 1d-maze is a simple problem from \citeA{Cassandra94actingoptimally}.
The agent begins at a random, non-goal location within a $1 \times 4$ maze.
There is a choice of two actions: left or right.
Each action transfers the agent to the adjacent cell if it exists, otherwise it has no effect.
If the agent reaches the third cell from the left, it receives a reward of $1$.
Otherwise it receives a reward of $0$.
The distinguishing feature of this problem is that the observations are \emph{uninformative};
every observation is the same regardless of the agent's actual location.

\paradot{Cheese Maze}
This well known problem is due to \citeA{mccallum96}.
The agent is a mouse inside a two dimensional maze seeking a piece of cheese.
The agent has to choose one of four actions: move up, down, left or right.
If the agent bumps into a wall, it receives a penalty of $-10$.
If the agent finds the cheese, it receives a reward of $10$.
Each movement into a free cell gives a penalty of $-1$.
The problem is depicted graphically in Figure \ref{fig:cheese_maze}.
The number in each cell represents the decimal equivalent of the four bit binary observation (0 for a free neighbouring cell, 1 for a wall) the mouse receives in each cell.
The problem exhibits perceptual aliasing in that a single observation is potentially ambiguous.

\begin{figure}[!h]
\centering
\includegraphics[width=16em]{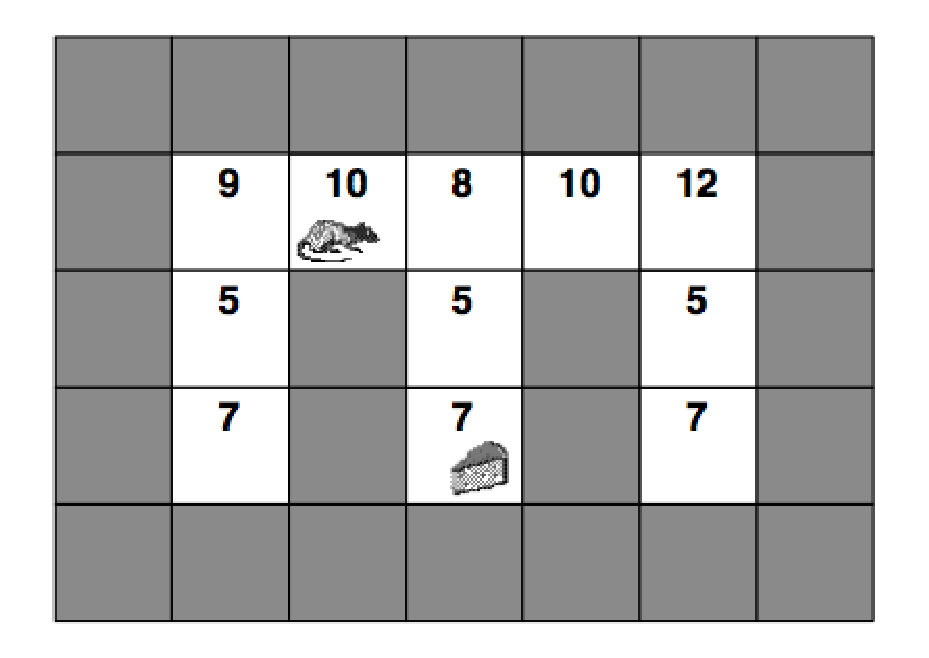}
\vspace{-1em}
\caption{The cheese maze}
\label{fig:cheese_maze}
\end{figure}

\paradot{Tiger}
This is another familiar domain from \citeA{Kaelbling95planningand}.
The environment dynamics are as follows: a tiger and a pot of gold are hidden behind one of two doors.
Initially the agent starts facing both doors. The agent has a choice of one of three actions: listen, open the left door, or open the right door.
If the agent opens the door hiding the tiger, it suffers a -100 penalty.
If it opens the door with the pot of gold, it receives a reward of 10.
If the agent performs the listen action, it receives a penalty of $-1$ and an observation that correctly describes where the tiger is with $0.85$ probability.

\paradot{Extended Tiger}
The problem setting is similar to Tiger, except that now the agent begins sitting down on a chair.
The actions available to the agent are: stand, listen, open the left door, and open the right door.
Before an agent can successfully open one of the two doors, it must stand up.
However, the listen action only provides information about the tiger's whereabouts when the agent is sitting down.
Thus it is necessary for the agent to plan a more intricate series of actions before it sees the optimal solution.
The reward structure is slightly modified from the simple Tiger problem, as now the agent gets a reward of 30 when finding the pot of gold.

\paradot{4 $\times$ 4 Grid}
The agent is restricted to a 4 $\times$ 4 grid world.
It can move either up, down, right or left.
If the agent moves into the bottom right corner, it receives a reward of $1$, and it is randomly teleported to one of the remaining $15$ cells.
If it moves into any cell other than the bottom right corner cell, it receives a reward of $0$.
If the agent attempts to move into a non-existent cell, it remains in the same location.
Like the 1d-maze, this problem is also uninformative but on a much larger scale.
Although this domain is simple, it does require some subtlety on the part of the agent.
The correct action depends on what the agent has tried before at previous time steps.
For example, if the agent has repeatedly moved right and not received a positive reward, then the chances of it receiving a positive reward by moving down are increased.

\paradot{TicTacToe}
In this domain, the agent plays repeated games of TicTacToe against an opponent who moves randomly.
If the agent wins the game, it receives a reward of $2$.
If there is a draw, the agent receives a reward of $1$.
A loss penalises the agent by $-2$.
If the agent makes an illegal move, by moving on top of an already filled square, then it receives a reward of $-3$.
A legal move that does not end the game earns no reward.

\paradot{Biased Rock-Paper-Scissors}
This domain is taken from \citeA{farias07}.
The agent repeatedly plays Rock-Paper-Scissor against an opponent that has a slight, predictable bias in its strategy.
If the opponent has won a round by playing rock on the previous cycle, it will always play rock at the next cycle; otherwise it will pick an action uniformly at random.
The agent's observation is the most recently chosen action of the opponent.
It receives a reward of $1$ for a win, $0$ for a draw and $-1$ for a loss.

\paradot{Kuhn Poker}
Our next domain involves playing Kuhn Poker \cite{kuhn50,hoehn05} against an opponent playing a Nash strategy.
Kuhn Poker is a simplified, zero-sum, two player poker variant that uses a deck of three cards: a King, Queen and Jack.
Whilst considerably less sophisticated than popular poker variants such as Texas Hold'em, well-known strategic concepts such as bluffing and slow-playing remain characteristic of strong play.

In our setup, the agent acts second in a series of rounds.
Two actions, pass or bet, are available to each player.
A bet action requires the player to put an extra chip into play.
At the beginning of each round, each player puts a chip into play.
The opponent then decides whether to pass or bet; betting will win the round if the agent subsequently passes, otherwise a showdown will occur.
In a showdown, the player with the highest card wins the round.
If the opponent passes, the agent can either bet or pass; passing leads immediately to a showdown, whilst betting requires the opponent to either bet to force a showdown, or to pass and let the agent win the round uncontested.
The winner of the round gains a reward equal to the total chips in play, the loser receives a penalty equal to the number of chips they put into play this round.
At the end of the round, all chips are removed from play and another round begins.

Kuhn Poker has a known optimal solution.
Against a first player playing a Nash strategy, the second player can obtain at most an average reward of $\frac{1}{18}$ per round.

\paradot{Partially Observable Pacman}
This domain is a partially observable version of the classic Pacman game.
The agent must navigate a $17 \times 17$ maze and eat the pills that are distributed across the maze.
Four ghosts roam the maze.
They move initially at random, until there is a Manhattan distance of 5 between them and Pacman, whereupon they will aggressively pursue Pacman for a short duration.
The maze structure and game are the same as the original arcade game, however the Pacman agent is hampered by partial observability.
Pacman is unaware of the maze structure and only receives a 4-bit observation describing the wall configuration at its current location.
It also does not know the exact location of the ghosts, receiving only 4-bit observations indicating whether a ghost is visible (via
direct line of sight) in each of the four cardinal directions.
In addition, the locations of the food pellets are unknown except for a 3-bit observation that indicates whether food can be smelt within
a Manhattan distance of 2, 3 or 4 from Pacman's location, and another 4-bit observation indicating whether there is food in its direct line of sight.
A final single bit indicates whether Pacman is under the effects of a power pill.
At the start of each episode, a food pellet is placed down with probability $0.5$ at every empty location on the grid.
The agent receives a penalty of 1 for each movement action, a penalty of 10 for running into a wall, a reward of $10$ for each food pellet eaten, a penalty of $50$ if it is caught by a ghost, and a reward of $100$ for collecting all the food.
If multiple such events occur, then the total reward is cumulative, i.e. running into a wall and being caught would give a penalty of $60$.
The episode resets if the agent is caught or if it collects all the food.

\begin{figure}[h]
\centering
\includegraphics[scale=0.22]{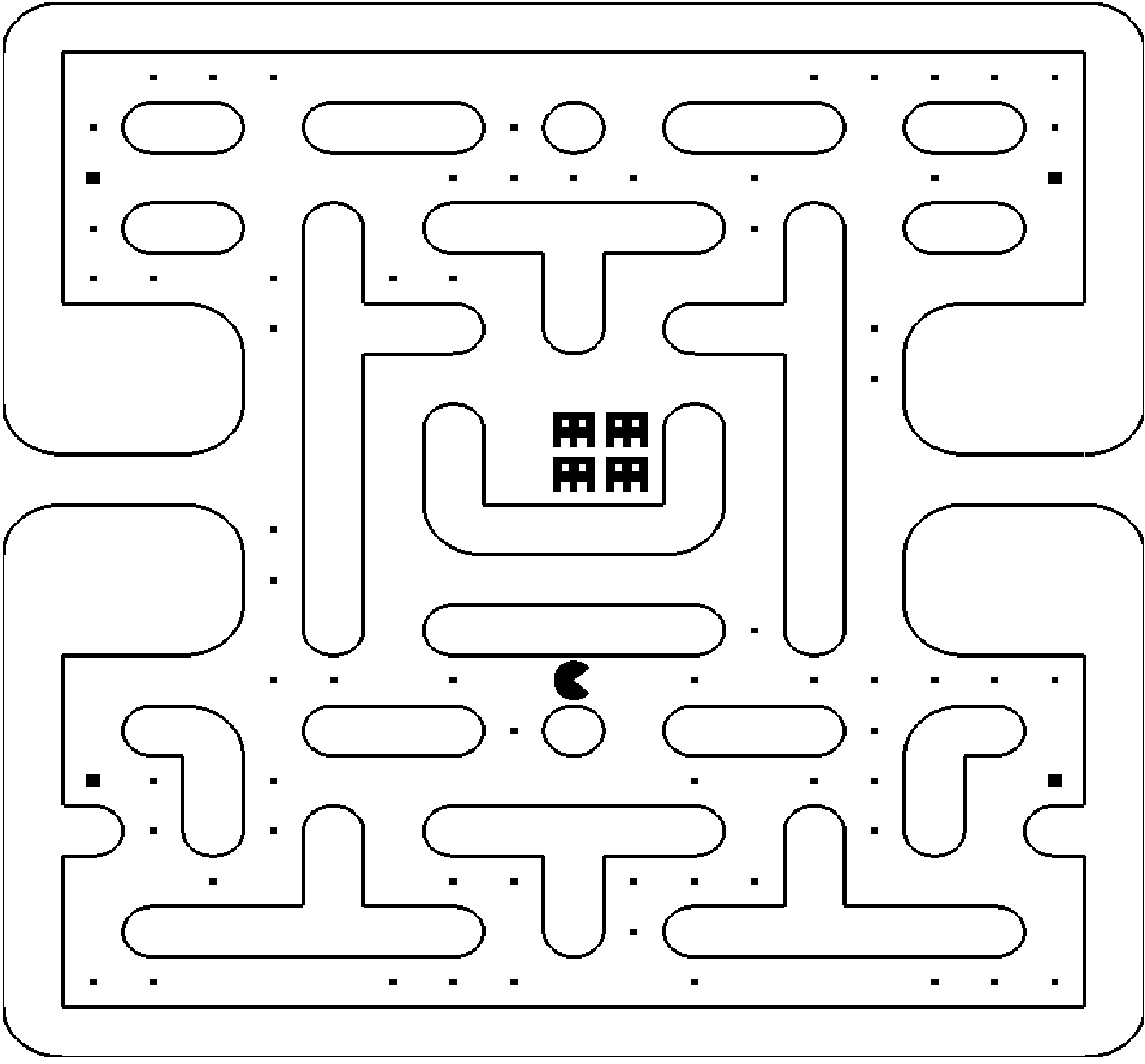}
\caption{A screenshot (converted to black and white) of the PacMan domain}
\label{fig:pocman}
\end{figure}

Figure \ref{fig:pocman} shows a graphical representation of the partially observable Pacman domain.
This problem is the largest domain we consider, with an unknown optimal policy.
The main purpose of this domain is to show the scaling properties of our agent on a challenging problem.
Note that this domain is fundamentally different to the Pacman domain used in \cite{silver10}.
In addition to using a different observation space, we also do not assume that the true environment is known a-priori.

\paradot{Experimental Setup}
We now evaluate the performance of the \agent\ agent.
To help put our results into perspective, we implemented and directly compared against two competing algorithms from the model-based general reinforcement learning literature: U-Tree \cite{mccallum96} and Active-LZ \cite{farias07}.
The two algorithms are described on page \pageref{u-tree description} in Section~\ref{sec:discussion}.
As \predictor\ subsumes Action Conditional CTW, we do not evaluate it in this paper; earlier results using Action Conditional CTW can be found in \cite{veness10}.
The performance of the agent using \predictor\ is no worse and in some cases slightly better than the previous results.

Each agent communicates with the environment over a binary channel.
A cycle begins with the agent sending an action $a$ to the environment, which then responds with a percept $x$.
This cycle is then repeated.
A fixed number of bits are used to encode the action, observation and reward spaces for each domain.
These are specified in Table \ref{table:domain_encoding}.
No constraint is placed on how the agent interprets the observation component; e.g., this could be done at either the bit or symbol level.
The rewards are encoded naively, i.e. the bits corresponding to the reward are interpreted as unsigned integers.
Negative rewards are handled (without loss of generality) by offsetting all of the rewards so that they are guaranteed to be non-negative.
These offsets are removed from the reported results.

\begin{table}[h]
\vspace{1.2em}
\begin{center}
\begin{tabular}{|l|c|c|c|}
\hline
Domain & $\cA$ bits & $\cO$ bits & $\cR$ bits \\
\hline
1d-maze & 1 & 1 & 1\\
Cheese Maze & 2 & 4 & 5\\
Tiger & 2 & 2 & 7\\
Extended Tiger & 2 & 3 & 8\\
4 $\times$ 4 Grid & 2 & 1 & 1\\
TicTacToe & 4 & 18 & 3\\
Biased Rock-Paper-Scissor & 2 & 2 & 2\\
Kuhn Poker & 1 & 4 & 3\\
Partially Observable Pacman & 2 & 16 & 8\\
\hline
\end{tabular}
\caption{Binary encoding of the domains}
\label{table:domain_encoding}
\end{center}
\end{table}

\begin{table}
\begin{center}
\begin{tabular}{|l|c|c|c|c|c|}
\hline
Domain & $D$ & $m$ & $\epsilon$ & $\gamma$ & \searchalg\ Simulations\\
\hline
1d-maze & 32 & 10 & 0.9 & 0.99 & 500\\
Cheese Maze & 96 & 8 & 0.999 & 0.9999 & 500\\
Tiger & 96 & 5 & 0.99 & 0.9999 & 500\\
Extended Tiger & 96 & 4 & 0.99 & 0.99999 & 500\\
4 $\times$ 4 Grid & 96 & 12 & 0.9 & 0.9999 & 500\\
TicTacToe & 64 & 9 & 0.9999 & 0.999999 & 500\\
Biased Rock-Paper-Scissor & 32 & 4 & 0.999 & 0.99999 & 500\\
Kuhn Poker & 42 & 2 & 0.99 & 0.9999 & 500\\
Partial Observable Pacman & 96 & 4 & 0.9999 & 0.99999 & 500\\
\hline
\end{tabular}
\caption{\agent\ model learning configuration}
\label{table:domain_description}
\end{center}
\vspace{-1em}
\end{table}

The process of gathering results for each of the three agents is broken into two phases: model learning and model evaluation.
The model learning phase involves running each agent with an exploratory policy to build a model of the environment.
This learnt model is then evaluated at various points in time by running the agent without exploration for $5000$ cycles and reporting the average reward per cycle.
More precisely, at time $t$ the average reward per cycle is defined as $\frac{1}{5000}\sum_{i=t+1}^{t+5000} r_i$, where $r_i$ is the reward received at cycle $i$.
Having two separate phases reduces the influence of the agent's earlier exploratory actions on the reported performance.
All of our experiments were performed on a dual quad-core Intel 2.53Ghz Xeon with 24 gigabytes of memory.

Table \ref{table:domain_description} outlines the parameters used by \agent\ during the model learning phase.
The context depth parameter $D$ specifies the maximal number of recent bits used by \predictor.
The \searchalg\ search horizon is specified by the parameter $m$.
Larger $D$ and $m$ increase the capabilities of our agent, at the expense of linearly increasing computation time; our values represent an appropriate compromise between these two competing dimensions for each problem domain.
Exploration during the model learning phase is controlled by the $\epsilon$ and $\gamma$ parameters.
At time $t$, \agent\ explores a random action with probability $\gamma^t\epsilon$.
During the model evaluation phase, exploration is disabled, with results being recorded for varying amounts of experience and search effort.

The Active-LZ algorithm is fully specified in \cite{farias07}.
It contains only two parameters, a discount rate and a policy that balances between exploration and exploitation.
During the model learning phase, a discount rate of $0.99$ and $\epsilon$-Greedy exploration (with $\epsilon=0.95$) were used.
Smaller exploration values (such as $0.05$, $0.2$, $0.5$) were tried, as well as policies that decayed $\epsilon$ over time, but these surprisingly gave slightly worse performance during testing.
As a sanity check, we confirmed that our implementation could reproduce the experimental results reported in \cite{farias07}.
During the model evaluation phase, exploration is disabled.

The situation is somewhat more complicated for U-Tree, as it is more of a general agent framework than a completely specified algorithm.
Due to the absence of a publicly available reference implementation, a number of implementation-specific decisions were made.
These included the choice of splitting criteria, how far back in time these criteria could be applied, the frequency of fringe tests, the choice of p-value for the Kolmogorov-Smirnov test, the exploration/exploitation policy and the learning rate.
The main design decisions are listed below:
\vspace{-0.5em}
\begin{itemize}\itemsep1mm\parskip0mm
\item A split could be made on any action, or on the status of any single bit of an observation.
\item The maximum number of steps backwards in time for which a utile distinction could be made was set to 5.
\item The frequency of fringe tests was maximised given realistic resource constraints. Our choices allowed for $5\times10^4$ cycles of interaction to be completed on each domain within 2 days of training time.
\item Splits were tried in order from the most temporally recent to the most temporally distant.
\item $\epsilon$-Greedy exploration strategy was used, with $\epsilon$ tuned separately for each domain.
\item The learning rate $\alpha$ was tuned for each domain.
\end{itemize}
To help make the comparison as fair as possible, an effort was made to tune U-Tree's parameters for each domain.
The final choices for the model learning phase are summarised in Table \ref{table:utree_parameters}.
During the model evaluation phase, both exploration and testing of the fringe are disabled.

\begin{table}
\begin{center}
\begin{tabular}{|l|c|c|c|}
\hline
Domain & $\epsilon$ & Test Fringe & $\alpha$ \\
\hline
1d-maze & 0.05 & 100 & 0.05\\
Cheese Maze & 0.2 & 100 & 0.05\\
Tiger & 0.1 & 100 & 0.05\\
Extended Tiger & 0.05 & 200 & 0.01\\
4 $\times$ 4 Grid & 0.05 & 100 & 0.05\\
TicTacToe & 0.05 & 1000 & 0.01\\
Biased Rock-Paper-Scissor & 0.05 & 100 & 0.05\\
Kuhn Poker & 0.05 & 200 & 0.05\\
\hline
\end{tabular}
\caption{U-Tree model learning configuration}
\label{table:utree_parameters}
\end{center}
\vspace{-1em}
\end{table}

\paradot{Source Code}
The code for our U-Tree, Active-LZ and \agent\ implementations can be found at:
\begin{small}\url{http://jveness.info/software/mcaixi_jair_2010.zip}\end{small}.

\paradot{Results}
Figure \ref{fig:reward_vs_age} presents our main set of results.
Each graph shows the performance of each agent as it accumulates more experience.
The performance of \agent\ matches or exceeds U-Tree and Active-LZ on all of our test domains.
Active-LZ steadily improved with more experience, however it learnt significantly more slowly than both U-Tree and \agent.
U-Tree performed well in most domains, however the overhead of testing for splits limited its ability to be run for long periods of time.
This is the reason why some data points for U-Tree are missing from the graphs in Figure \ref{fig:reward_vs_age}.
This highlights the advantage of algorithms that take constant time per cycle, such as \agent\ and Active-LZ.
Constant time isn't enough however, especially when large observation spaces are involved.
Active-LZ works at the symbol level, with the algorithm given by \citeA{farias07} requiring an exhaustive enumeration of the percept space on each cycle.
This is not possible in reasonable time for the larger TicTacToe domain, which is why no Active-LZ result is presented.
This illustrates an important advantage of \agent\ and U-Tree, which have the ability to exploit structure \emph{within} a single observation.

\begin{figure}
\vspace{-7em}
\centering
\subfigure{
\vspace{-5.5em}
}
\vspace{-5em}
\subfigure{
\hspace{-5.65em}
\includegraphics[scale=0.365]{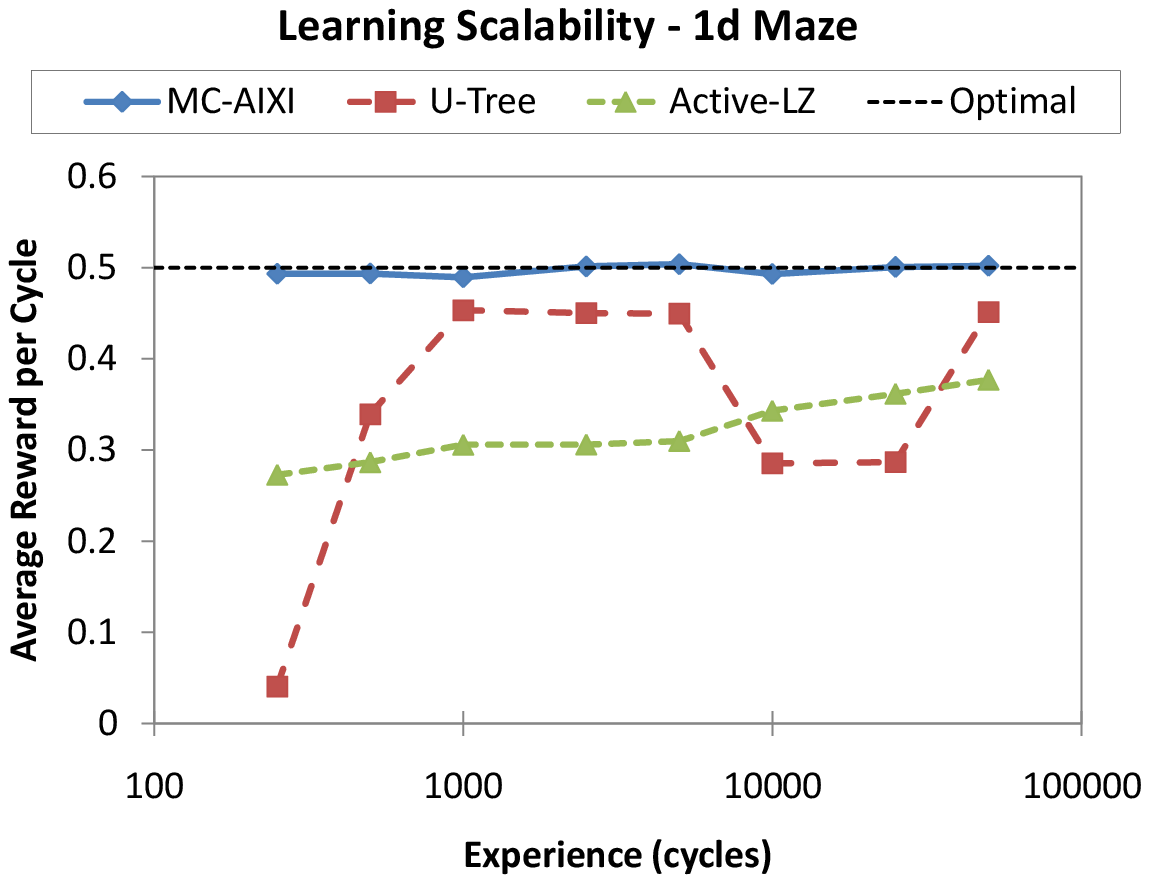}
\hspace{-8em}
\includegraphics[scale=0.365]{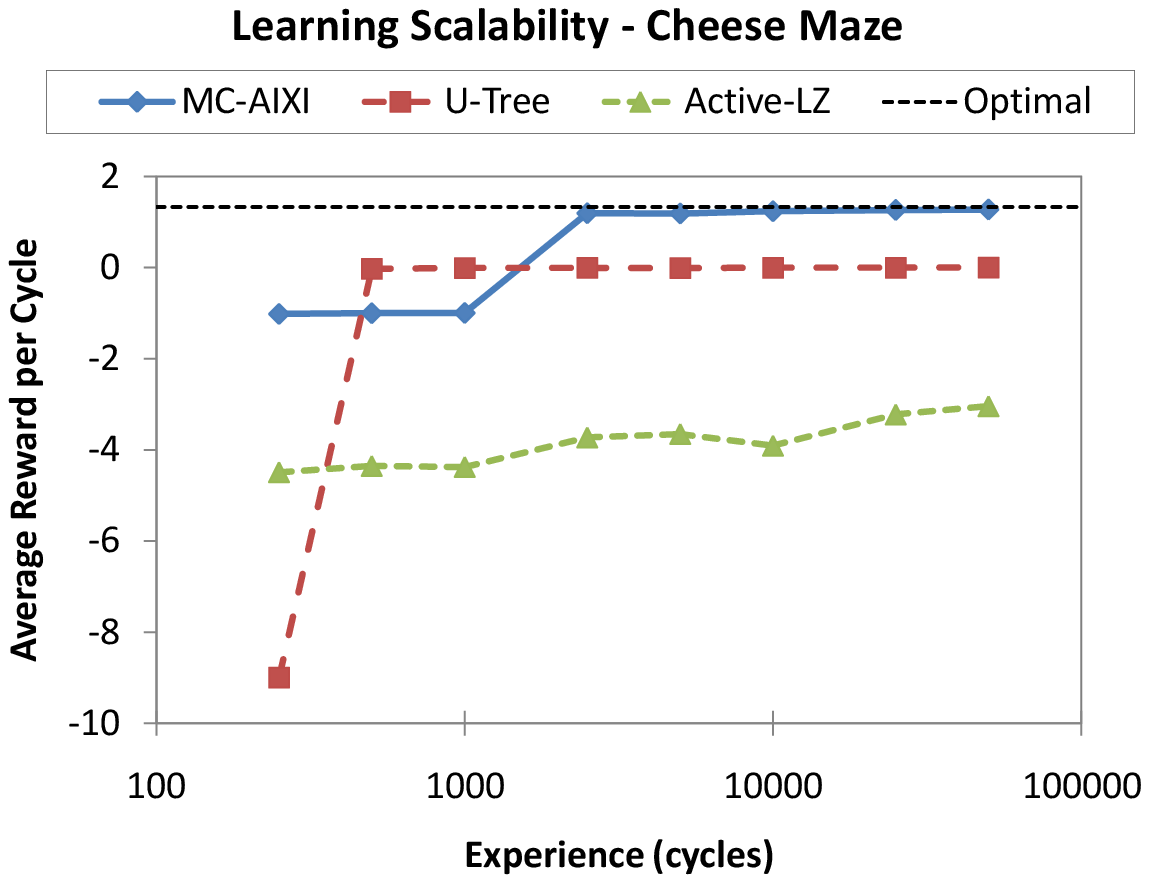}
}
\vspace{-5.5em}
\subfigure{
\hspace{-5.65em}
\includegraphics[scale=0.365]{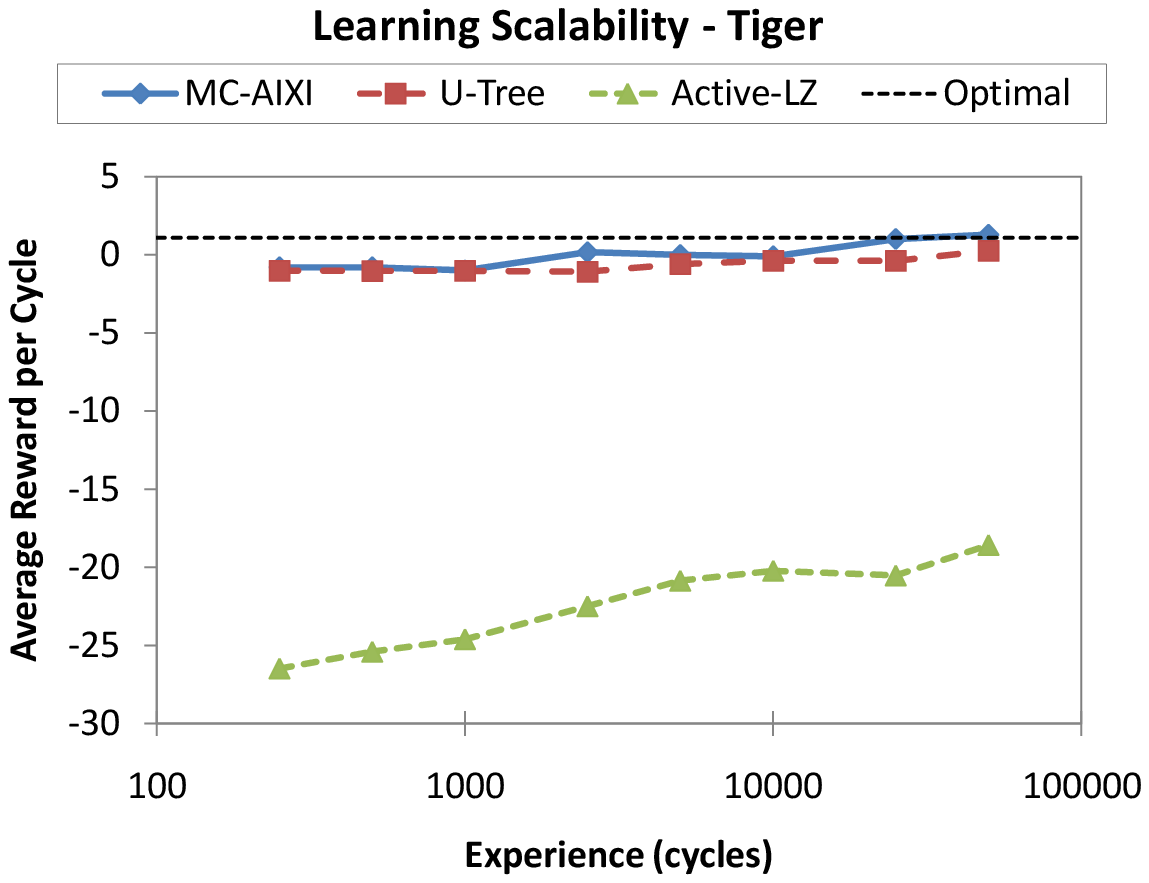}
\hspace{-8em}
\includegraphics[scale=0.365]{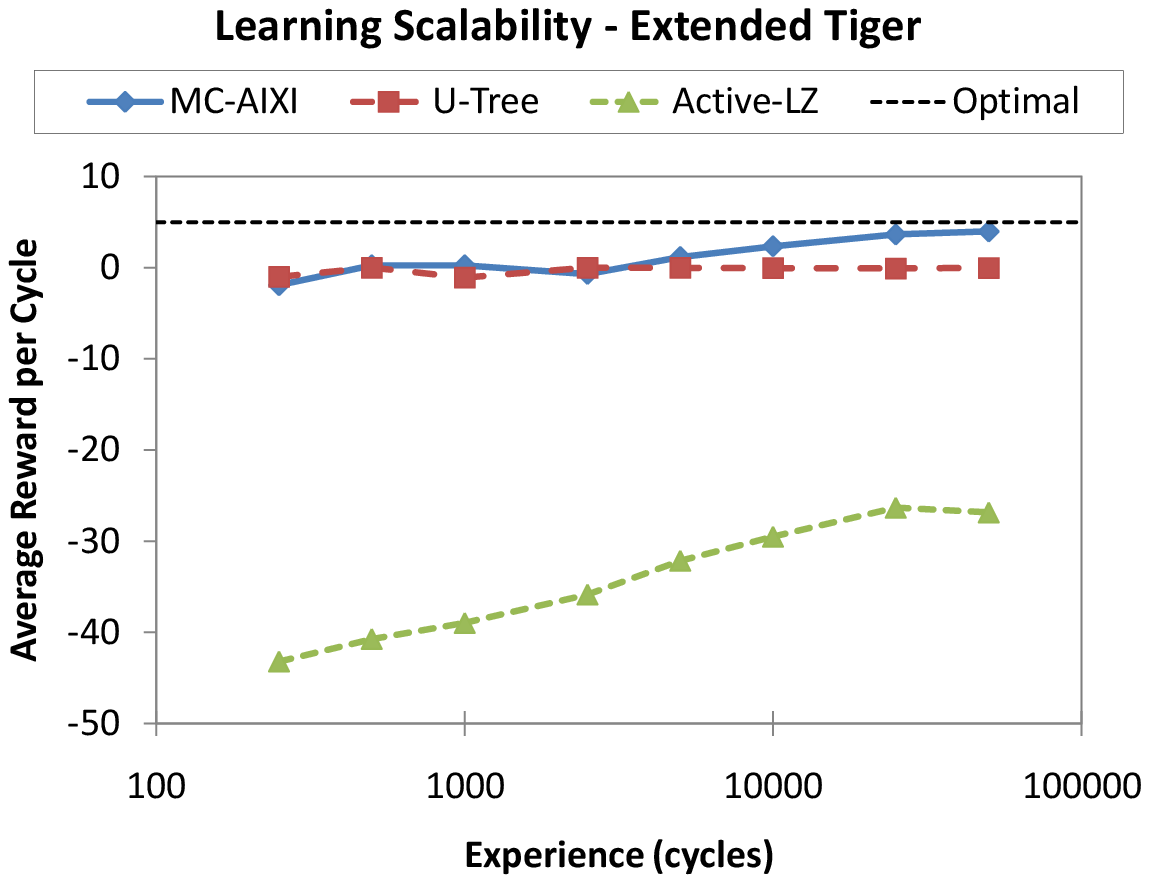}
}
\vspace{-5.5em}
\subfigure{
\hspace{-5.65em}
\includegraphics[scale=0.365]{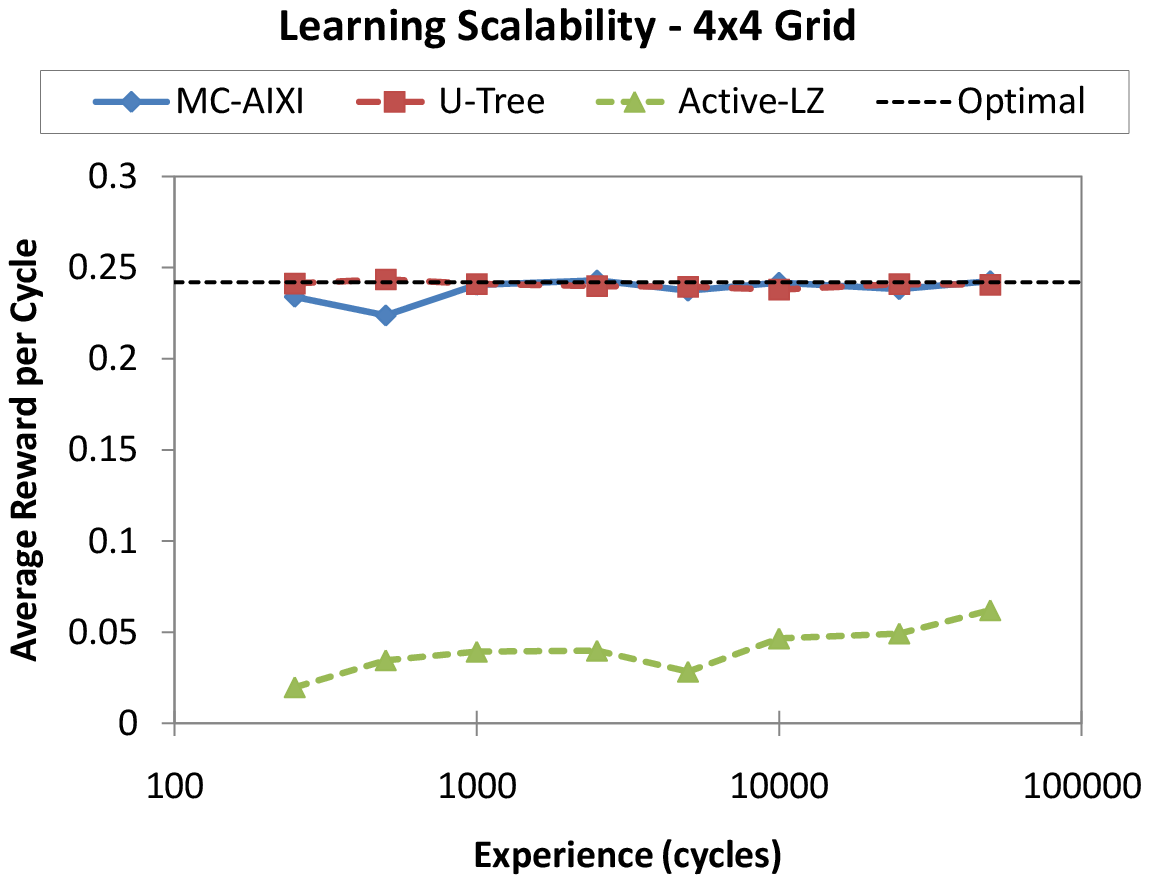}
\hspace{-8em}
\includegraphics[scale=0.365]{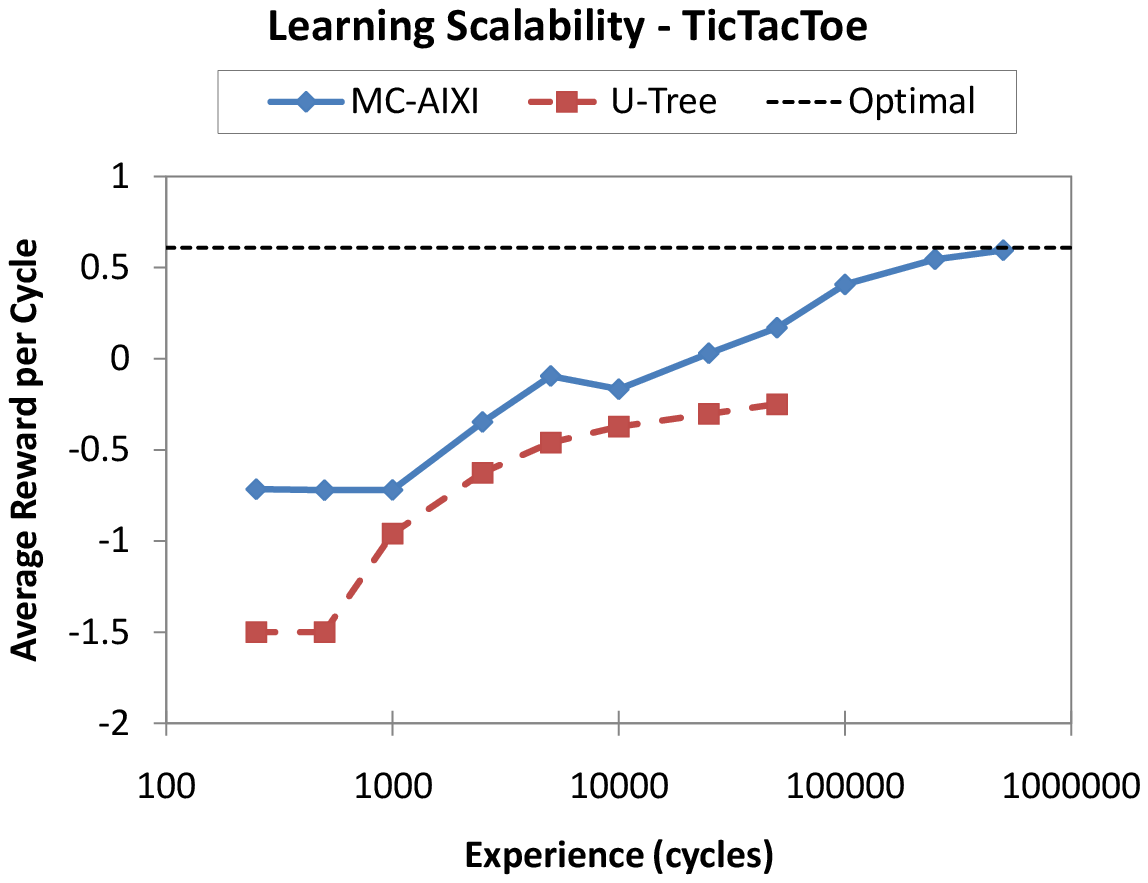}
}
\vspace{-5.5em}
\subfigure{
\hspace{-5.65em}
\includegraphics[scale=0.365]{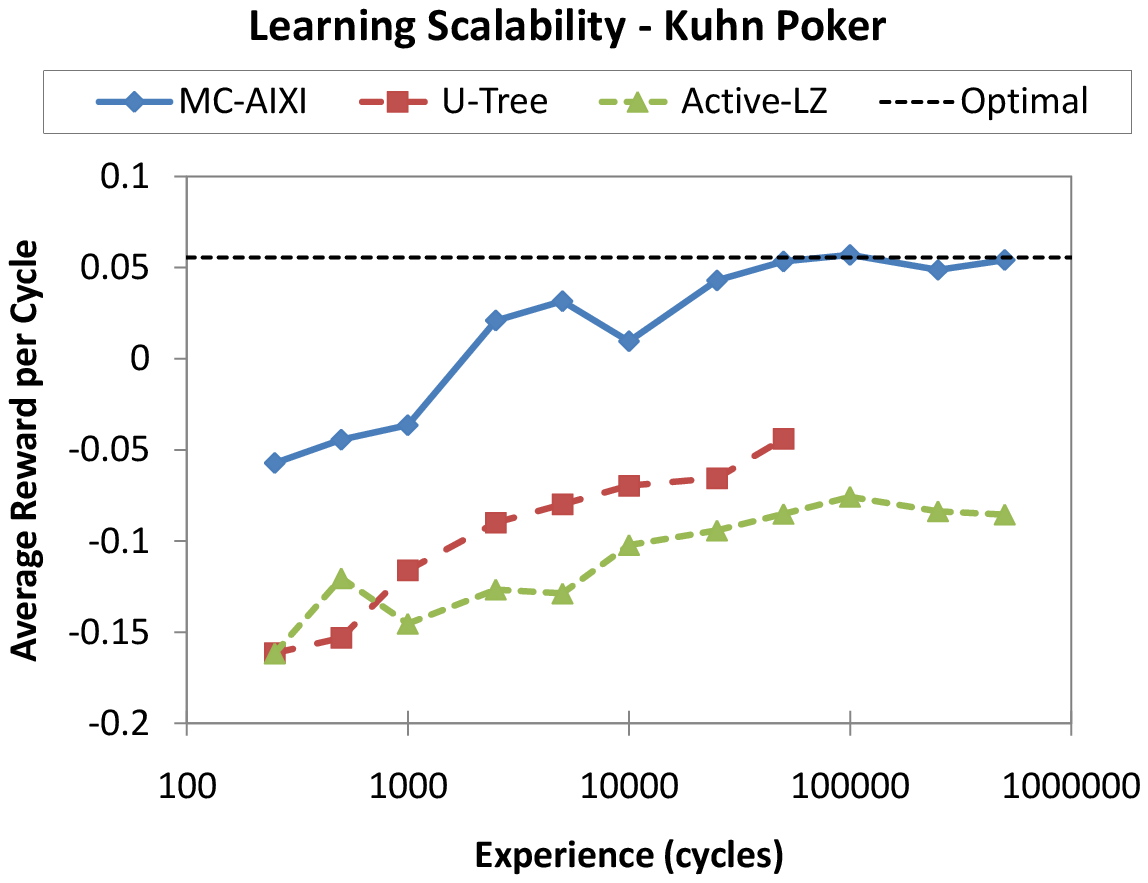}
\hspace{-8em}
\includegraphics[scale=0.365]{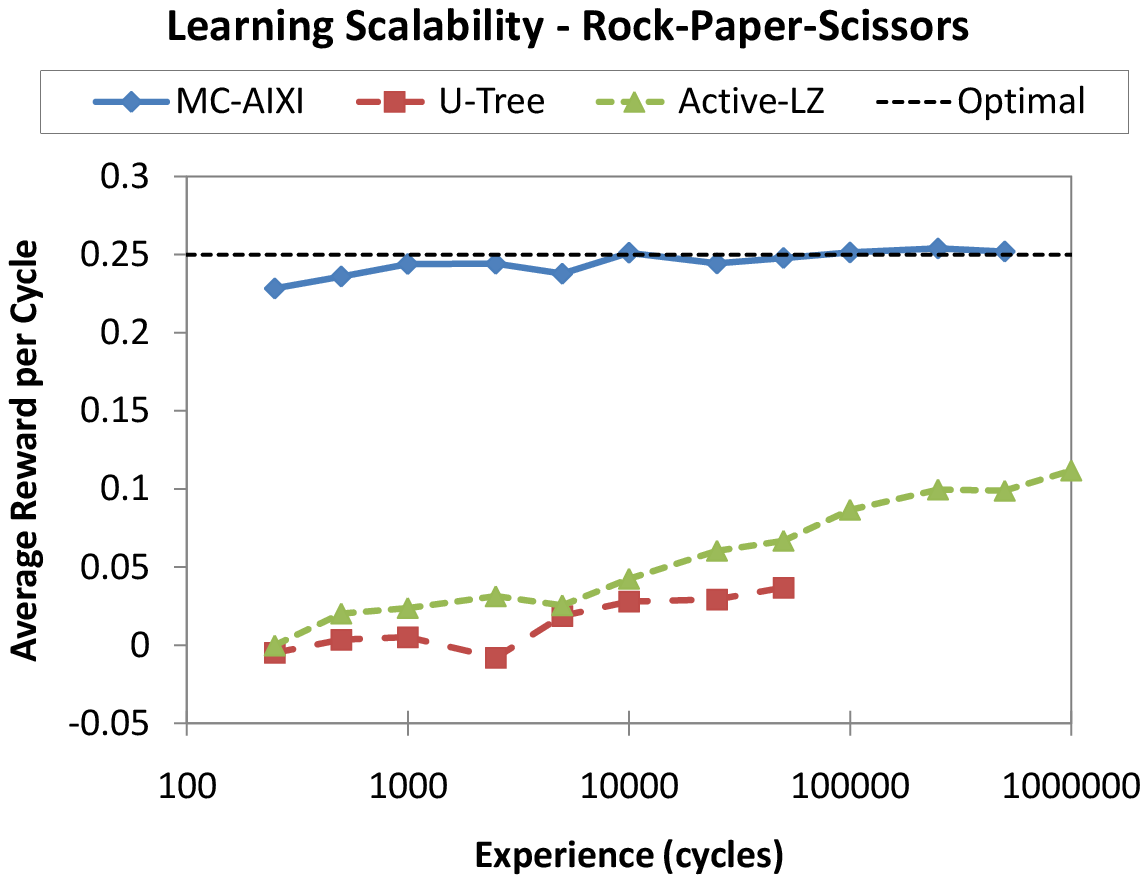}
}
\vspace{1.8em}
\caption{Average Reward per Cycle vs Experience}
\label{fig:reward_vs_age}
\end{figure}

\begin{figure}[t!]
\vspace{-5em}
\centerline{
\includegraphics[scale=0.58]{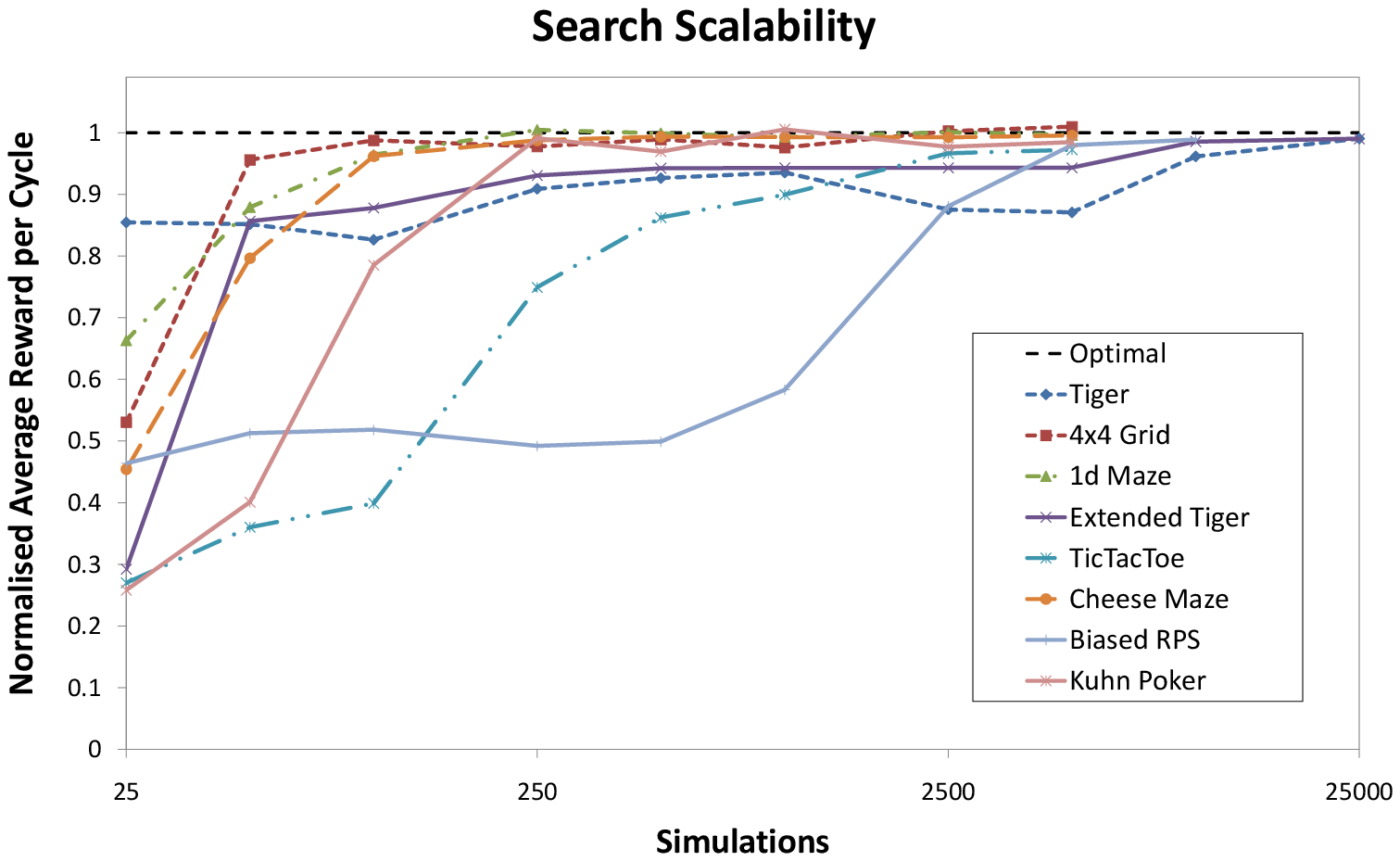}
}
\vspace{-4.5em}
\caption{Performance versus \searchalg\ search effort}
\label{fig:reward_vs_time}
\end{figure}

Figure \ref{fig:reward_vs_time} shows the performance of \agent\ as the number of \searchalg\ simulations varies.
The results for each domain were based on a model learnt from $5 \times 10^4$ cycles of experience, except in the case of TicTacToe where $5\times 10^5$ cycles were used.
So that results could be compared across domains, the average reward per cycle was normalised to the interval $[0,1]$.
As expected, domains that included a significant planning component (such as Tiger or Extended Tiger) required more search effort.
Good performance on most domains was obtained using only 1000 simulations.

\begin{table}
\vspace{0.5em}
\begin{center}
\begin{tabular}{|l|c|c|c|}
\hline
 Domain &  Experience &  \searchalg\ Simulations &  Search Time per Cycle\\
\hline
 1d Maze & $5 \times 10^3$ & 250 & 0.1s\\
 Cheese Maze &  $2.5 \times 10^3$ &  500 &  0.5s\\
 Tiger &  $2.5 \times 10^4$ &  25000 &  10.6s\\
 Extended Tiger & $5 \times 10^4$ & 25000 & 12.6s\\
 4 $\times$ 4 Grid &  $2.5 \times 10^4$ &  500 &  0.3s\\
 TicTacToe &  $5 \times 10^5$ &  2500 &  4.1s\\
 Biased RPS &  $1 \times 10^4$ &  5000 &  2.5s \\
 Kuhn Poker &  $5 \times 10^6$ &  250 &  0.1s\\
\hline
\end{tabular}
\caption{Resources required for (near) optimal performance by MC-AIXI(\sc fac-ctw)}
\label{table:results}
\end{center}
\vspace{-1.5em}
\end{table}

Given a sufficient number of \searchalg\ simulations and cycles of interaction, the performance of the \agent\ agent approaches optimality on our test domains.
The amount of resources needed for near optimal performance on each domain during the model evaluation phase is listed in Table \ref{table:results}.
Search times are also reported.
This shows that the \agent\ agent can be realistically used on a present day workstation.

\paradot{Discussion}
The small state space induced by U-Tree has the benefit of limiting the number of parameters that need to be estimated from data.
This can dramatically speed up the model-learning process.
In contrast, both Active-LZ and our approach require a number of parameters proportional to the number of distinct contexts.
This is one of the reasons why Active-LZ exhibits slow convergence in practice.
This problem is much less pronounced in our approach for two reasons.
First, the Ockham prior in CTW ensures that future predictions are dominated by PST structures that have seen enough data to be trustworthy.
Secondly, value function estimation is decoupled from the process of context estimation.
Thus it is reasonable to expect \searchalg\ to make good local decisions provided \predictor\ can predict well.
The downside however is that our approach requires search for action selection.
Although \searchalg\ is an anytime algorithm, in practice more computation (at least on small domains) is required per cycle compared to approaches like Active-LZ and U-Tree that act greedily with respect to an estimated global value function.

The U-Tree algorithm is well motivated, but unlike Active-LZ and our approach, it lacks theoretical performance guarantees.
It is possible for U-Tree to prematurely converge to a locally optimal state representation from which the heuristic splitting criterion can never recover.
Furthermore, the splitting heuristic contains a number of configuration options that can dramatically influence its performance \cite{mccallum96}.
This parameter sensitivity somewhat limits the algorithm's applicability to the general reinforcement learning problem.
Still, our results suggest that further investigation of frameworks motivated along the same lines as U-Tree is warranted.

\paradot{Comparison to 1-ply Rollout Planning}
We now investigate the performance of \searchalg\ in comparison to an adaptation of the well-known 1-ply rollout-based planning technique of \citeA{bertsekas99}.
In our setting, this works as follows:
given a history $h$, an estimate $\hat{V}(ha)$ is constructed for each action $a \in \cA$, by averaging the returns of many length $m$ simulations initiated from $ha$.
The first action of each simulation is sampled uniformly at random from $\cA$, whilst the remaining actions are selected according to some heuristic rollout policy.
Once a sufficient number of simulations have been completed, the action with the highest estimated value is selected.
Unlike \searchalg, this procedure doesn't build a tree, nor is it guaranteed to converge to the depth $m$ expectimax solution.
In practice however, especially in noisy and highly stochastic domains, rollout-based planning can significantly improve the performance of an existing heuristic rollout policy \cite{bertsekas99}.

Table \ref{table:mc_vs_mcts} shows how the performance (given by average reward per cycle) differs when \searchalg\ is replaced by the 1-ply rollout planner.
The amount of experience collected by the agent, as well as the total number of rollout simulations, is the same as in Table \ref{table:results}.
Both \searchalg\ and the 1-ply planner use the same search horizon, heuristic rollout policy (each action is chosen uniformly at random) and total number of simulations for each decision.
This is reasonable, since although \searchalg\ has a slightly higher overhead compared to the 1-ply rollout planner, this difference is negligible when taking into account the cost of simulating future trajectories using \predictor.
Also, similar to previous experiments, $5000$ cycles of greedy action selection were used to evaluate the performance of the \predictor\ + 1-ply rollout planning combination.

\begin{table}[h!]
\vspace{0.5em}
\begin{center}
\begin{tabular}{|l|c|c|}
\hline
 Domain &  \agent\ &  \predictor\ + 1-ply MC \\
\hline
1d Maze       &  0.50   &  0.50\\
Cheese Maze    &  1.28    &  1.25\\
Tiger          &  1.12    &  1.11\\
\textbf{Extended Tiger} &  \textbf{3.97}  & \textbf{-0.97}\\
4x4 Grid       &  0.24    &  0.24\\
TicTacToe      &  0.60    &  0.59\\
\textbf{Biased RPS}     &  \textbf{0.25}    &  \textbf{0.20}\\
Kuhn Poker     &  0.06 &  0.06\\
\hline
\end{tabular}
\caption{Average reward per cycle: \searchalg\ versus 1-ply rollout planning}
\label{table:mc_vs_mcts}
\end{center}
\vspace{-1.5em}
\end{table}

Importantly, \searchalg\ never gives worse performance than the 1-ply rollout planner, and on some domains (shown in bold) performs better.
The \searchalg\ algorithm provides a way of performing multi-step planning whilst retaining the considerable computational advantages of rollout based methods.
In particular, \searchalg\ will be able to construct deep plans in regions of the search space where most of the probability mass is concentrated on a small set of the possible percepts.
When such structure exists, \searchalg\ will automatically exploit it.
In the worst case where the environment is highly noisy or stochastic, the performance will be similar to that of rollout based planning.
Interestingly, on many domains the empirical performance of 1-ply rollout planning matched that of \searchalg.
We believe this to be a byproduct of our modest set of test domains, where multi-step planning is less important than learning an accurate model of the environment.

\paradot{Performance on a Challenging Domain}
The performance of \agent\ was also evaluated on the challenging Partially Observable Pacman domain.
This is an enormous problem.
Even if the true environment were known, planning would still be difficult due to the $10^{60}$ distinct underlying states.

We first evaluated the performance of \agent\ online.
A discounted $\epsilon$-Greedy policy, which chose a random action at time $t$ with probability $\epsilon\gamma^t$ was used.
These parameters were instantiated with $\epsilon:=0.9999$ and $\gamma:=0.99999$.
When not exploring, each action was determined by \searchalg\ using $500$ simulations.
Figure \ref{fig:pacman-online} shows both the average reward per cycle and the average reward across the most recent 5000 cycles.

The performance of this learnt model was then evaluated by performing $5000$ steps of greedy action selection, at various time points, whilst varying the number of simulations used by \searchalg.
Figure \ref{fig:pacman-scaling} shows obtained results.
The agent's performance scales with both the number of cycles of interaction and the amount of search effort.
The results in Figure \ref{fig:pacman-scaling} using $500$ simulations are higher than in Figure \ref{fig:pacman-online} since the performance is no longer affected by the exploration policy or earlier behavior based on an inferior learnt model.

\begin{figure}[t]
\vspace{-5em}
\centerline{
\includegraphics[scale=0.55]{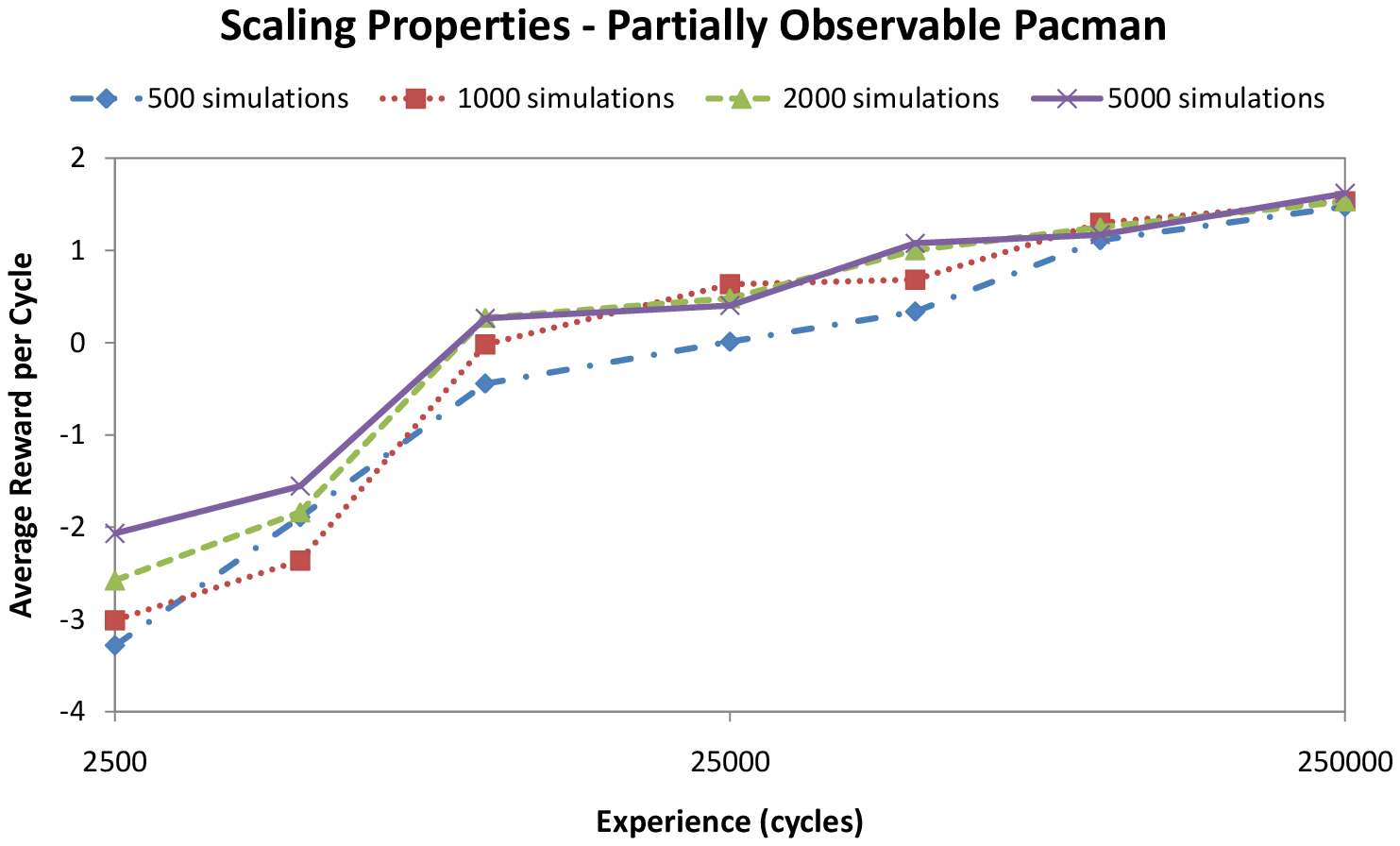}
}
\vspace{-5em}
\caption{Scaling properties on a challenging domain}
\label{fig:pacman-scaling}
\end{figure}

Visual inspection\footnote{See \begin{scriptsize}\url{http://jveness.info/publications/pacman_jair_2010.wmv}\end{scriptsize} for a graphical demonstration} of Pacman shows that the agent, whilst not playing perfectly, has already learnt a number of important concepts.
It knows not to run into walls.
It knows how to seek out food from the limited information provided by its sensors.
It knows how to run away and avoid chasing ghosts.
The main subtlety that it hasn't learnt yet is to aggressively chase down ghosts when it has eaten a red power pill.
Also, its behaviour can sometimes become temporarily erratic when stuck in a long corridor with no nearby food or visible ghosts.
Still, the ability to perform reasonably on a large domain and exhibit consistent improvements makes us optimistic about the ability of the \agent\ agent to scale with extra computational resources.

\section{Discussion}\label{sec:discussion}

\paradot{Related Work}
There have been several attempts at studying the computational properties of AIXI.
In \citeA{hutter02fastest}, an asymptotically optimal algorithm is proposed that, in parallel, picks and runs the fastest program from an enumeration
of provably correct programs for any given well-defined problem.
A similar construction that runs all programs of length less than $l$ and time less than $t$ per cycle and picks the best output (in the
sense of maximising a provable lower bound for the true value) results in the optimal time bounded AIXI$tl$ agent \cite[Chp.7]{Hutter:04uaibook}.
Like Levin search \cite{levin73}, such algorithms are not practical in general but can in some cases be applied
successfully; see e.g. \citeA{schmidhuber97a,schmidhuber97b,schmidhuber03,schmidhuber04}.
In tiny domains, universal learning is computationally feasible with brute-force search.
In \cite{poland06},  the behaviour of AIXI is compared with a universal predicting-with-expert-advice algorithm \cite{poland05} in repeated $2\times 2$ matrix
games and is shown to exhibit different behaviour.
A Monte-Carlo algorithm is proposed by \citeA{pankov08} that samples programs according to their algorithmic probability as a way of
approximating Solomonoff's universal prior.
A closely related algorithm is that of speed prior sampling \cite{schmidhuber02}.

\begin{figure}[t!]
\vspace{-8.2em}
\centerline{\includegraphics[scale=0.55]{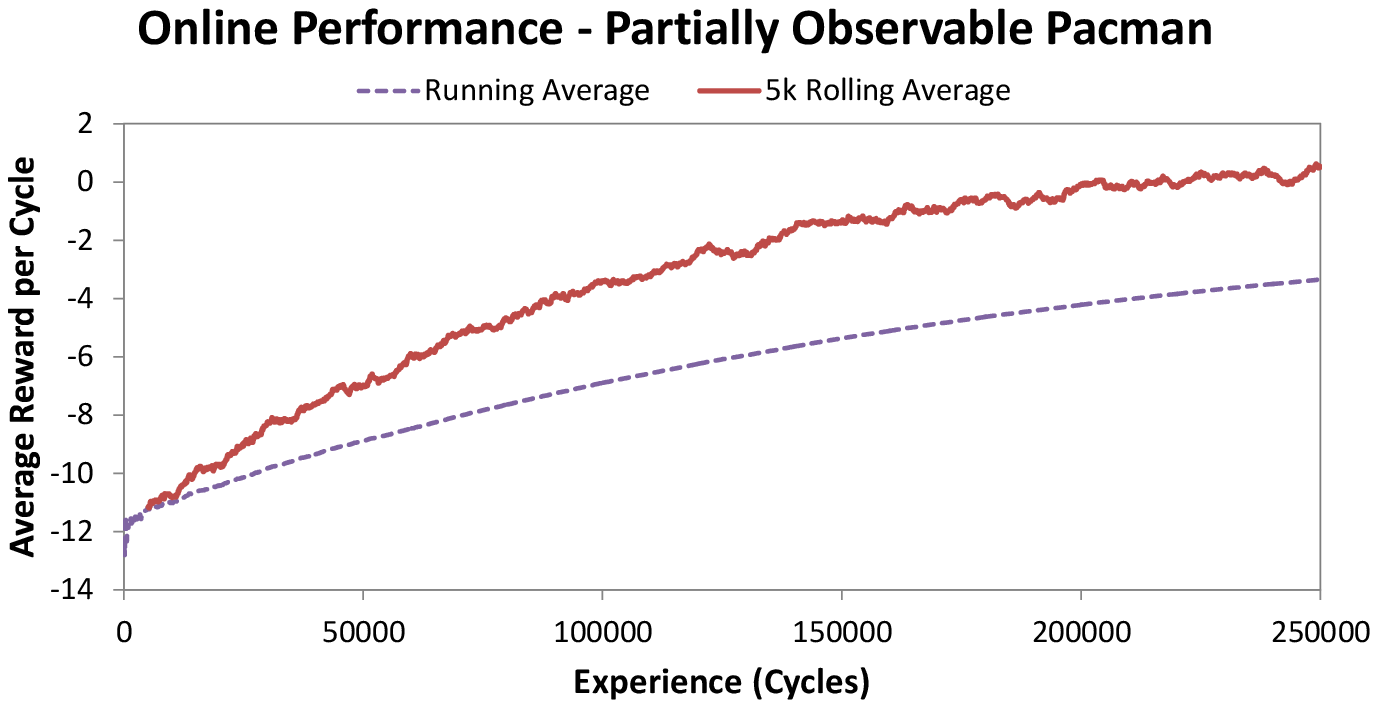}}
\vspace{-6.5em}
\caption{Online performance on a challenging domain}
\label{fig:pacman-online}
\end{figure}

We now move on to a discussion of the model-based general reinforcement learning literature.
An early and influential work is the Utile Suffix Memory (USM) algorithm described by \citeA{mccallum96}.
USM uses a suffix tree to partition the agent's history space into distinct states, one for each leaf in the suffix tree.
Associated with each state/leaf is a Q-value, which is updated incrementally from experience like in Q-learning \cite{watkins92}.
The history-partitioning suffix tree is grown in an incremental fashion, starting from a single leaf node in the beginning.
A leaf in the suffix tree is split when the history sequences that fall into the leaf are shown to exhibit statistically different Q-values.
The USM algorithm works well for a number of tasks but could not deal effectively with noisy environments.
Several extensions of USM to deal with noisy environments are investigated in \cite{shani04,shani07}.

\label{u-tree description}
U-Tree \cite{mccallum96} is an online agent algorithm that attempts to discover a compact state representation from a raw stream of experience.
The main difference between U-Tree and USM is that U-Tree can discriminate between individual components within an observation.
This allows U-Tree to more effectively handle larger observation spaces and ignore potentially irrelevant components of the observation vector.
Each state is represented as the leaf of a suffix tree that maps history sequences to states.
As more experience is gathered, the state representation is refined according to a heuristic built around the Kolmogorov-Smirnov test.
This heuristic tries to limit the growth of the suffix tree to places that would allow for better prediction of future reward.
Value Iteration is used at each time step to update the value function for the learnt state representation, which is then used by the agent for action selection.

Active-LZ \cite{farias07} combines a Lempel-Ziv based prediction scheme with dynamic programming for control to produce an agent that is provably asymptotically optimal if the environment is $n$-Markov.
The algorithm builds a context tree (distinct from the context tree built by CTW), with each node containing accumulated transition statistics and a value function estimate.
These estimates are refined over time, allowing for the Active-LZ agent to steadily increase its performance.
In Section \ref{sec:experiments}, we showed that our agent compared favourably to Active-LZ.

The BLHT algorithm \cite{suematsu97,suematsu99} uses symbol level PSTs for learning and an (unspecified) dynamic programming based algorithm for control.
BLHT uses the most probable model for prediction, whereas we use a mixture model, which admits a much stronger convergence result.
A further distinction is our usage of an Ockham prior instead of a uniform prior over PST models.

Predictive state representations (PSRs) \cite{psr02,psr04,rosencrantz2004lld} maintain predictions of future experience.
Formally, a PSR is a probability distribution over the agent's future experience, given its past experience.
A subset of these predictions, the core tests, provide a sufficient statistic for all future experience.
PSRs provide a Markov state representation, can represent and track the agent's state in partially observable environments, and provide a complete model of the world's dynamics.
Unfortunately, exact representations of state are impractical in large domains, and some form of approximation is typically required.
Topics such as improved learning or discovery algorithms for PSRs are currently active areas of research.
The recent results of \citeA{boots10} appear particularly promising.

Temporal-difference networks \cite{SuttonT04} are a form of predictive state representation in which the agent's state is approximated by abstract predictions. These can be predictions about future observations, but also predictions about future predictions. This set of interconnected predictions is known as the {\em question network}. Temporal-difference networks learn an approximate model of the world's dynamics: given the current predictions, the agent's action, and an observation vector, they provide new predictions for the next time-step. The parameters of the model, known as the {\em answer network}, are updated after each time-step by temporal-difference learning.
Some promising recent results applying TD-Networks for prediction (but not control) to small POMDPs are given in \cite{makino09}.

In model-based Bayesian Reinforcement Learning \cite{strens00,pascal06,ross07,pascal08}, a distribution over (PO)MDP parameters is maintained.
In contrast, we maintain an exact Bayesian mixture of PSTs, which are variable-order Markov models.
The \searchalg\ algorithm shares similarities with Bayesian Sparse Sampling \cite{wang05}.
The main differences are estimating the leaf node values with a rollout function and using the UCB policy to direct the search.

\paradot{Limitations}
Our current AIXI approximation has two main limitations.

The first limitation is the restricted model class used for learning and prediction.
Our agent will perform poorly if the underlying environment cannot be predicted well by a PST of bounded depth.
Prohibitive amounts of experience will be required if a large PST model is needed for accurate prediction.
For example, it would be unrealistic to think that our current AIXI approximation could cope with real-world image or audio data.

The second limitation is that unless the planning horizon is unrealistically small, our full Bayesian solution (using \searchalg\ and a mixture environment model) to the exploration/exploitation dilemma is computationally intractable.
This is why our agent needs to be augmented by a heuristic exploration/exploitation policy in practice.
Although this did not prevent our agent from obtaining optimal performance on our test domains, a better solution may be required for more challenging problems.
In the MDP setting, considerable progress has been made towards resolving the exploration/exploitation issue.
In particular, powerful PAC-MDP approaches exist for both model-based and model-free reinforcement learning agents \cite{brafman02,strehl06,strehl09}.
It remains to be seen whether similar such principled approaches exist for history-based Bayesian agents.

\section{Future Scalability}\label{sec:future_work}

We now list some ideas that make us optimistic about the future scalability of our approach.

\paradot{Online Learning of Rollout Policies for \searchalg}
An important parameter to \searchalg\ is the choice of rollout policy.
In MCTS methods for Computer Go, it is well known that search performance can be improved by using knowledge-based rollout policies \cite{gelly2006tr}.
In the general agent setting, it would thus be desirable to gain some of the benefits of expert design through online learning.

We have conducted some preliminary experiments in this area.
A CTW-based method was used to predict the high-level actions chosen online by \searchalg.
This learnt distribution replaced our previous uniformly random rollout policy.
Figure \ref{fig:bootstrapped_playouts_cheese} shows the results of using this learnt rollout policy on the cheese maze.
The other domains we tested exhibited similar behaviour.
Although more work remains, it is clear that even our current simple learning scheme can significantly improve the performance of \searchalg.

Although our first attempts have been promising, a more thorough investigation is required.
It is likely that rollout policy learning methods for adversarial games, such as \cite{silver09}, can be adapted to our setting.
It would also be interesting to try to apply some form of search bootstrapping \cite{veness09} online.
In addition, one could also look at ways to modify the UCB policy used in \searchalg\ to automatically take advantage of learnt rollout knowledge, similar to the heuristic techniques used in computer Go \cite{gelly07}.

\begin{figure}
\vspace{-4.5em}
\centerline{\includegraphics[scale=0.5]{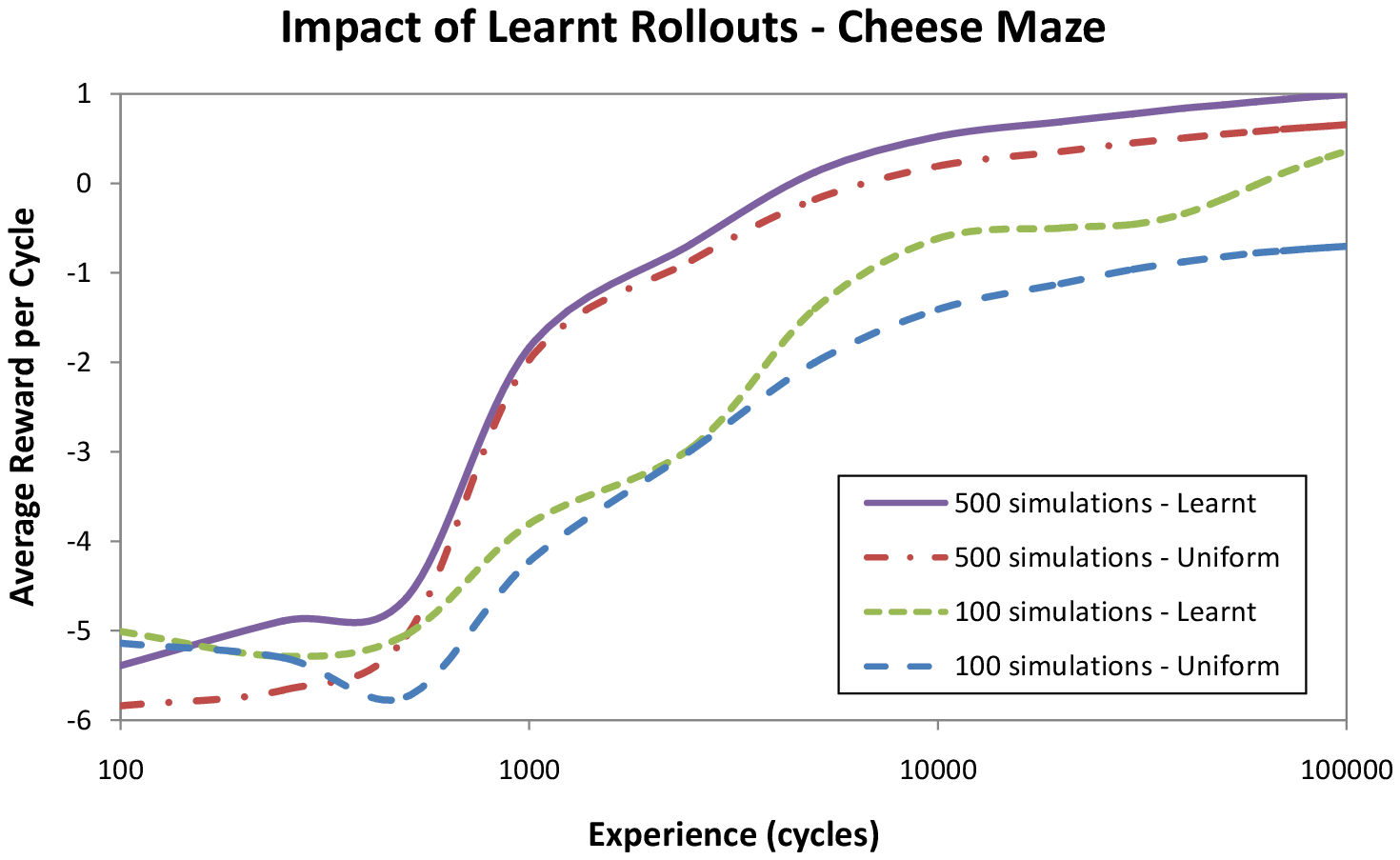}}
\vspace{-4.5em}
\caption{Online performance when using a learnt rollout policy on the Cheese Maze}
\label{fig:bootstrapped_playouts_cheese}
\end{figure}

\paradot{Combining Mixture Environment Models}
A key property of mixture environment models is that they can be \emph{composed}.
Given two mixture environment models $\xi_1$ and $\xi_2$, over model classes $\mathcal{M}_1$ and $\mathcal{M}_2$ respectively, it is easy to show that the convex combination
\begin{equation*}
\xi(x_{1:n} \cbar a_{1:n}) := \alpha \xi_1(x_{1:n} \cbar a_{1:n}) + (1 - \alpha) \xi_2(x_{1:n} \cbar a_{1:n})
\end{equation*}
is a mixture environment model over the union of $\mathcal{M}_1$ and $\mathcal{M}_2$.
Thus there is a principled way for expanding the general predictive power of agents that use our kind of direct AIXI approximation.

\paradot{Richer Notions of Context for FAC-CTW}\label{subsec:predicate ctw}
Instead of using the most recent $D$ bits of the current history $h$, the FAC-CTW algorithm can be generalised to use a set of $D$ boolean functions on $h$ to define the current context.
We now formalise this notion, and give some examples of how this might help in agent applications.

\begin{defn}
Let ${\cal P} = \{ p_0, p_1, \ldots, p_m \}$ be a set of predicates (boolean functions) on histories $h \in ( {\cal A} \times {\cal X} )^n, n \geq 0$.
A $\cP$-model is a binary tree where each internal node is labeled with a predicate in $\cP$ and the left and right outgoing edges at the node
are labeled True and False respectively.
A $\cP$-tree is a pair $(M_\cP,\Theta)$ where $M_\cP$ is a $\cP$-model and associated with each leaf node $l$ in $M_\cP$ is a probability
distribution over $\{0,1\}$ parametrised by $\theta_l \in \Theta$.
\end{defn}

A ${\cal P}$-tree $(M_\cP,\Theta)$ represents a function $g$ from histories to probability distributions on $\{0,1\}$ in the usual way.
For each history $h$, $g(h) = \theta_{l_h}$, where $l_h$ is the leaf node reached by pushing $h$ down the model $M_\cP$ according to whether
it satisfies the predicates at the internal nodes and $\theta_{l_h} \in \Theta$ is the distribution at $l_h$.
The notion of a $\cP$-context tree can now be specified, leading to a natural generalisation of Definition \ref{defn:context tree}.

Both the Action-Conditional CTW and FAC-CTW algorithms can be generalised to work with $\cP$-context trees in a natural way.
Importantly, a result analogous to Lemma~\ref{model averaging} can be established, which means that the desirable computational properties of CTW are retained.
This provides a powerful way of extending the notion of context for agent applications.
For example, with a suitable choice of predicate class $\cP$, both prediction suffix trees (Definition \ref{defn:pst}) and looping suffix trees \cite{holmes06} can be represented as $\cP$-trees.
It also opens up the possibility of using rich logical tree models \cite{blockeel98topdown,kramer-widmer01,LogicforLearning,ng05thesis,lloyd-ng-learnModal} in place of prediction suffix trees.

\paradot{Incorporating CTW Extensions}
There are several noteworthy ways the original CTW algorithm can be extended.
The finite depth limit on the context tree can be removed \cite{Willems94ctwext}, without increasing the asymptotic space overhead of the algorithm.
Although this increases the worst-case time complexity of generating a symbol from $O(D)$ to linear in the length of the history, the average-case performance may still be sufficient for good performance in the agent setting.
Furthermore, three additional model classes, each significantly larger than the one used by CTW, are presented in \cite{willems96gcw}.
These could be made action conditional along the same lines as our FAC-CTW derivation.
Unfortunately, online prediction with these more general classes is now exponential in the context depth $D$.
Investigating whether these ideas can be applied in a more restricted sense would be an interesting direction for future research.

\paradot{Parallelization of \searchalg}
The performance of our agent is dependent on the amount of thinking time allowed at each time step.
An important property of \searchalg\ is that it is naturally parallel.
We have completed a prototype parallel implementation of \searchalg\, with promising scaling results using between 4 and 8 processing cores.
We are confident that further improvements to our implementation will allow us to solve problems where our agent's planning ability is the main limitation.

\paradot{Predicting at Multiple Levels of Abstraction}
The FAC-CTW algorithm reduces the task of predicting a single percept to the prediction of its binary representation.
Whilst this is reasonable for a first attempt at AIXI approximation, it's worth emphasising that subsequent attempts need not work exclusively at such a low level.

For example, recall that the FAC-CTW algorithm was obtained by chaining together $l_X$ action-conditional binary predictors.
It would be straightforward to apply a similar technique to chain together multiple $k$-bit action-conditional predictors, for $k>1$.
These $k$ bits could be interpreted in many ways: e.g. integers, floating point numbers, ASCII characters or even pixels.
This observation, along with the convenient property that mixture environment models can be composed, opens up the possibility of constructing more sophisticated, \emph{hierarchical} mixture environment models.

\section{Conclusion}

This paper presents the first computationally feasible general reinforcement learning agent  that {\em directly} and {\em scalably} approximates the AIXI ideal.
Although well established theoretically, it has previously been unclear whether the AIXI theory could inspire the design of practical agent algorithms.
Our work answers this question in the affirmative: empirically, our approximation achieves strong performance and theoretically, we can characterise the range of environments in which our agent is expected to perform well.

To develop our approximation, we introduced two new algorithms: \searchalg, a Monte-Carlo expectimax approximation technique that can be used with any online Bayesian approach to the general reinforcement learning problem and {\sc FAC-CTW}, a generalisation of the powerful CTW algorithm to the agent setting.
In addition, we highlighted a number of interesting research directions that could improve the performance of our current agent; in particular, model class expansion and the online learning of heuristic rollout policies for \searchalg.

We hope that this work generates further interest from the broader artificial intelligence community in both the AIXI theory and general reinforcement learning agents.

\paradot{Acknowledgments}
The authors thank Alan Blair, Thomas Degris-Dard, Evan Greensmith, Bernhard Hengst, Ramana Kumar, John Lloyd, Hassan Mahmud, Malcolm Ryan, Scott Sanner, Rich Sutton, Eric Wiewiora, Frans Willems and the anonymous reviewers for helpful comments and feedback.
This work received support from the Australian Research Council under grant DP0988049.
NICTA is funded by the Australian Government's Department of Communications, Information Technology, and the Arts and the Australian Research Council through Backing Australia's Ability and the ICT Research Centre of Excellence programs.

\addcontentsline{toc}{section}{\refname}

\begin{small}
\newcommand{\etalchar}[1]{$^{#1}$}

\end{small}

\end{document}